\documentclass{article}

\usepackage{microtype}
\usepackage{graphicx}
\usepackage{booktabs} 

\usepackage{hyperref}

\usepackage[accepted]{icml2019}

\usepackage{amsfonts}
\usepackage{amsthm}
\usepackage{amsmath}
\usepackage{xspace}
\usepackage{graphicx}
\usepackage{subfig}

\newcommand{\mz}{\textsc{Montezuma's Revenge}\xspace}
\newcommand{\pitfall}{\textsc{Pitfall!}\xspace}
\newcommand{\pe}{\textsc{Private Eye}\xspace}
\newcommand{\ours}{\textsc{Absplore}\xspace}  \newcommand\abstraction{\phi}
\newcommand\knownset{\mathcal{S}}

\newcommand{\go}[2]{\text{go}(#1, #2)}

       \newcommand\refeqn[1]{(\ref{eqn:#1})}

\newcommand\refsec[1]{Section~\ref{sec:#1}}
\newcommand\refsubsec[1]{Section~\ref{subsec:#1}}

\newcommand\reffig[1]{Figure~\ref{fig:#1}}

\newcommand\reftab[1]{Table~\ref{tab:#1}}
\newcommand\refapp[1]{Appendix~\ref{sec:#1}}

\newcommand\reflem[1]{Lemma~\ref{lem:#1}}

\newcommand\refalg[1]{Algorithm~\ref{alg:#1}}

\newcommand\refprop[1]{Proposition~\ref{prop:#1}}

\ifthenelse{\isundefined{\definition}}{\newtheorem{definition}{Definition}}{}
\ifthenelse{\isundefined{\assumption}}{\newtheorem{assumption}{Assumption}}{}
\ifthenelse{\isundefined{\hypothesis}}{\newtheorem{hypothesis}{Hypothesis}}{}
\ifthenelse{\isundefined{\proposition}}{\newtheorem{proposition}{Proposition}}{}
\ifthenelse{\isundefined{\theorem}}{\newtheorem{theorem}{Theorem}}{}
\ifthenelse{\isundefined{\lemma}}{\newtheorem{lemma}{Lemma}}{}
\ifthenelse{\isundefined{\corollary}}{\newtheorem{corollary}{Corollary}}{}
\ifthenelse{\isundefined{\alg}}{\newtheorem{alg}{Algorithm}}{}
\ifthenelse{\isundefined{\example}}{\newtheorem{example}{Example}}{}

\usepackage{amsmath,amsfonts,bm}

\def\eqref#1{equation~\ref{#1}}

\def\1{\bm{1}}

\DeclareMathAlphabet{\mathsfit}{\encodingdefault}{\sfdefault}{m}{sl}
\SetMathAlphabet{\mathsfit}{bold}{\encodingdefault}{\sfdefault}{bx}{n}

\DeclareMathOperator*{\argmax}{arg\,max}

\icmltitlerunning{Learning Abstract Models for Strategic Exploration and Fast
Reward Transfer}

\begin{document}

\twocolumn[
\icmltitle{Learning Abstract Models for Strategic Exploration and Fast
Reward Transfer}

\icmlsetsymbol{resident}{$\dagger$}

\begin{icmlauthorlist}
\icmlauthor{Evan Zheran Liu}{cs,google,resident}
\icmlauthor{Ramtin Kermati}{cs}
\icmlauthor{Sudarshan Seshadri}{cs}
\icmlauthor{Kelvin Guu}{stats,google}
\icmlauthor{Panupong Pasupat}{cs,google}
\icmlauthor{Emma Brunskill}{cs}
\icmlauthor{Percy Liang}{cs,stats}
\end{icmlauthorlist}

\icmlaffiliation{cs}{Department of Computer Science, Stanford University,
Stanford, California}
\icmlaffiliation{stats}{Department of Statistics, Stanford University,
Stanford, California}
\icmlaffiliation{google}{Google Research, Mountain View, California, USA}

\icmlcorrespondingauthor{Evan Zheran Liu}{evanliu@cs.stanford.edu}

\icmlkeywords{Machine Learning, ICML}

\vskip 0.3in
]

\printAffiliationsAndNotice{\textsuperscript{$\dagger$}Work done as a member of the Google AI Residency
Program (\url{g.co/airesidency}) }

\begin{abstract}
  Model-based reinforcement learning (RL) is appealing because
  (i) it enables planning and thus more strategic exploration, and
  (ii) by decoupling dynamics from rewards, it enables fast
  transfer to new reward functions.
  However, learning an accurate Markov Decision Process (MDP) over
  high-dimensional states (e.g., raw pixels) is extremely challenging
  because it requires function approximation, which leads to compounding
  errors.
  Instead, to avoid compounding errors,
  we propose learning an \emph{abstract MDP} over abstract states:
  low-dimensional coarse representations of the state (e.g., capturing
  agent position, ignoring other objects).
We assume access to an abstraction function that maps the concrete states to
  abstract states.
  In our approach, we construct an abstract MDP,
  which grows through strategic exploration via planning.
  Similar to hierarchical RL approaches, the abstract actions of the
  abstract MDP are backed by learned subpolicies that navigate between abstract states.
Our approach achieves strong results on three of the hardest
  Arcade Learning Environment games (\mz, \pitfall, and \pe),
  including superhuman performance on \pitfall without demonstrations.
  After training on one task, we can reuse the learned abstract MDP
  for new reward functions, achieving higher reward in 1000x fewer samples
  than model-free methods trained from scratch.\footnote{
    Videos of our trained agent:
    \url{https://sites.google.com/view/abstract-models/home}}
\end{abstract}
 \section{Introduction}

Model-based reinforcement learning (RL) offers two advantages:
First, one can plan under a dynamics model, which produces a
strategic exporation policy \cite{brafman2002r},
which is crucial for high-dimensional, sparse reward settings.
Second, if we want to solve multiple tasks (e.g., loading or unloading the
dishwasher) that share the same dynamics (e.g., physics) but differ only in the
reward function, the dynamics learned from a \emph{single} task
can be ported directly to other tasks without additional learning
\cite{laroche2017transfer}.
However, learning accurate models in high-dimensional state spaces is
extremely challenging.
As one must resort to function approximation,
any errors in the model \emph{compound} \citep{talvitie2014model,
talvitie2015agnostic} across multiple timesteps when one tries to use the model for planning,
resulting in wildly inaccurate long-term
predictions \citep{oh2015action, finn2016unsupervised, chiappa2017recurrent}.

Inspired by state abstraction from hierarchical reinforcement learning
\cite{sutton1999between, dietterich2000state, andre2002state}
we propose a method, which we call Abstract Exploration (\ours), for learning an abstract MDP which avoids compounding errors
using a manager and worker.
We assume access to an abstraction function
\citep{li2006towards, singh1995reinforcement, dietterich1998maxq},
which maps a high-dimensional concrete state (e.g. all pixels on the screen) to a
low-dimensional abstract state (e.g. the position of the agent).
We aim to learn an (abstract) Markov Decision Process (MDP)
(\refsec{abstract_mdp}) over this abstract state space as follows:
A \emph{manager} maintains an abstract MDP over a subset of all possible
abstract states which we call the \emph{known set}, which is grown over time
(\refsec{growing}).
The crucial property we enforce is that this abstract MDP is highly accurate
and near deterministic on the known set, so we can revisit novel state regions
via planning while avoiding compounding errors.

Concurrently, we learn a \emph{worker policy} that the manager uses to
transition between abstract states (\refsec{worker}).
The worker policy operates on concrete states;
its goal is to hide the messy details of the real world from the manager (e.g., jumping over monsters)
so that the manager has a much simpler planning problem (e.g., traversing between two locations).
In our approach, the worker keeps an inventory of \emph{skills} (i.e., options \citep{sutton1999between}),
each of which is a neural network subpolicy;
the worker assigns an appropriate skill for each transition between abstract states.
In this way, the worker does not ``forget'' \citep{kirkpatrick2017overcoming},
and we ensure monotonic progress in learning the abstract MDP.

We evaluate the strategic exploration of \ours on the Arcade Learning
Environment (ALE) \citep{bellemare2013arcade} games with the sparsest rewards:
\mz, \pitfall, and \pe.
In \pitfall,
\ours achieves the superhuman performance without demonstrations.
It also achieves strong results on \mz and \pe,
although it is not directly comparable many prior approaches,
because it uses additional RAM state information in its abstraction function.
Additionally, \ours can use the learned dynamics of the abstract MDP
to transfer to new reward functions without re-training.
When evaluated on new reward functions never seen during training,
\ours achieves over 3x the reward in 1000x fewer samples than
model-free methods explicitly trained on the new rewards from scratch.

 \section{Abstract Markov Decision Processes}\label{sec:abstract_mdp}

\begin{figure}\centering
    \includegraphics[height=0.44\linewidth]{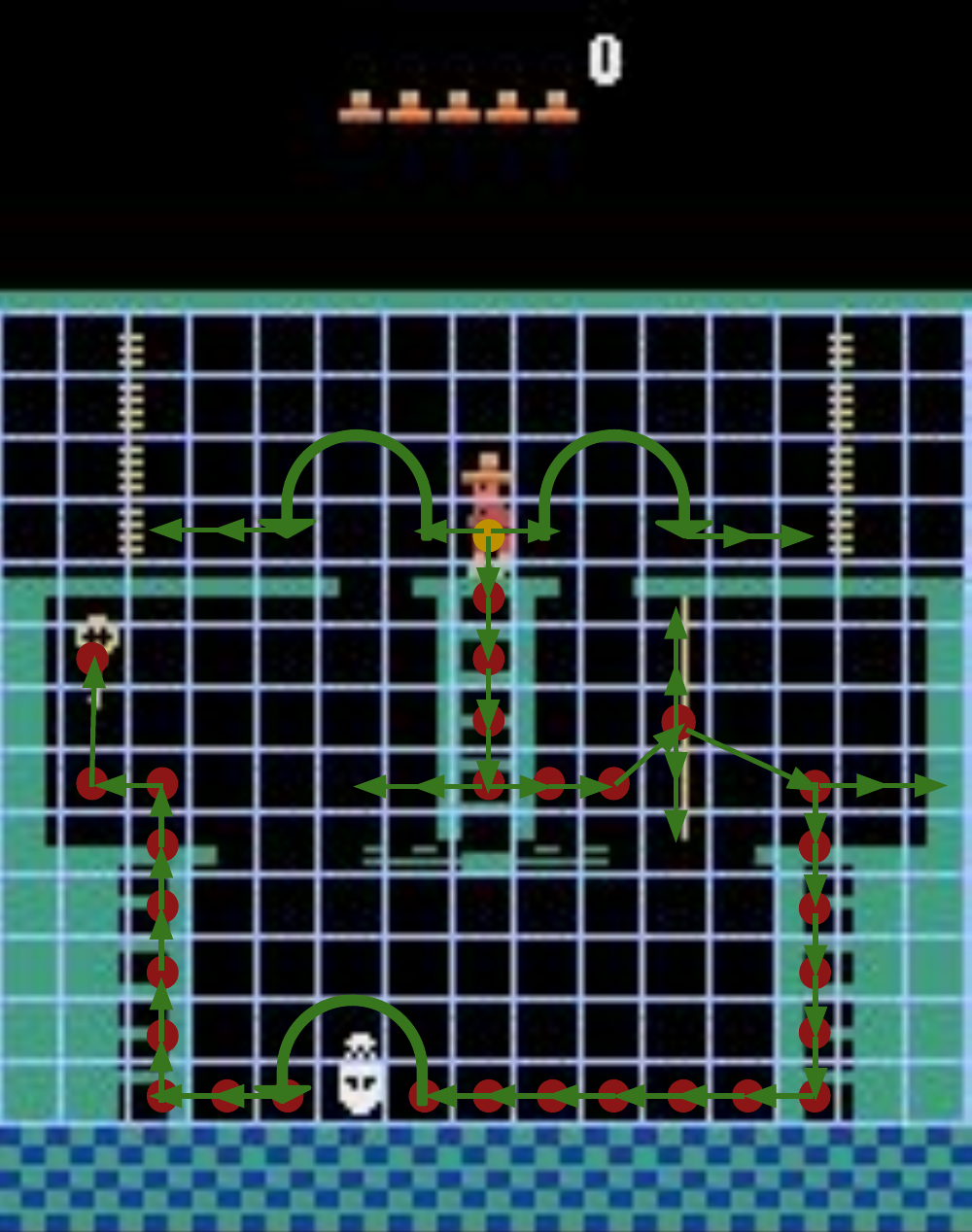}

    \caption{
      Illustration of the abstract MDP on \mz.
      We have superimposed a white grid on top of the original game.
      At any given time, the agent is in one of the grid cells -- each grid cell
      is an abstract state.
      In this example, the agent starts at the top of a ladder (yellow dot).
      The worker then navigates transitions (abstract actions) between
      abstract states (green arrows) to follow a plan made by the manager (red
      dots).
    }\label{fig:high_level_approach}\end{figure}

\begin{figure}
    \centering
    \subfloat[]{{\includegraphics[height=0.44\linewidth]{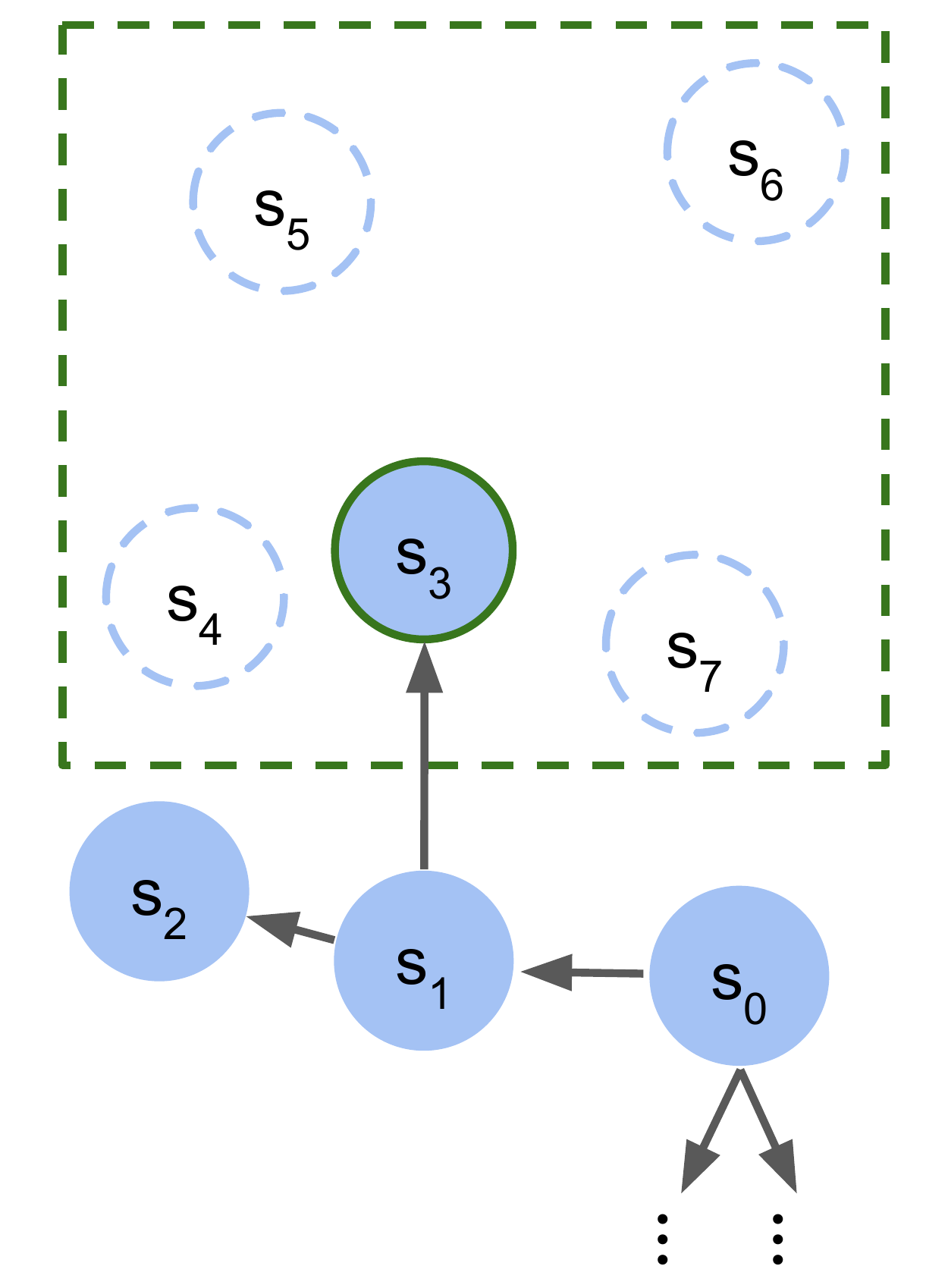}\hspace*{.5cm} }}
    \subfloat[]{{\includegraphics[height=0.44\linewidth]{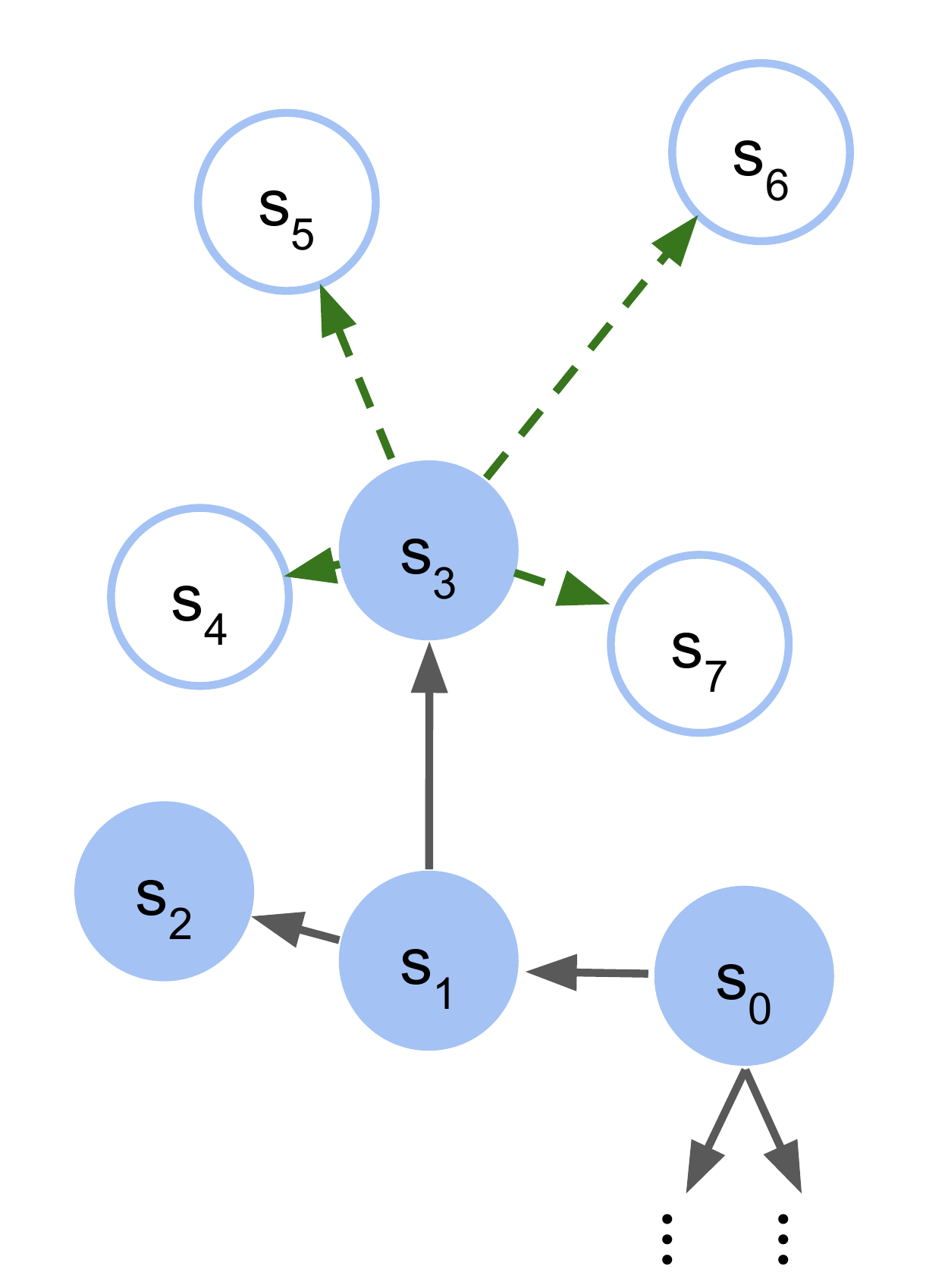} }}
    \caption{
      (a)
      Circles represent abstract states.
      Shaded circles represent states within the known set.
      The manager navigates the agent to the fringe of the known set ($s_3$),
      then randomly explores with $\pi^d$ to discover new candidate
      transitions near $s_3$ (dotted box).
(b) The worker extends the abstract MDP by learning to navigate the
      newly discovered candidate transitions (dotted arrows).
    }\label{fig:abstract_mdp}\end{figure}

We assume the world is an unknown episodic finite-horizon MDP with
(concrete) states $x \in \mathcal{X}$ and
actions $a \in \mathcal{A}$.
We further assume we have a simple predefined \emph{state abstraction function}
mapping concrete states $x$ to abstract states $s = \abstraction(x)$.
In \mz, for instance,
a concrete state contains the pixels on the screen,
while the corresponding abstract state contains the agent's
position and inventory (\reffig{high_level_approach}).
We assume that reward only depends on the abstract states:
taking any action transitioning from concrete state $x$ to $x'$ leads to
reward $R(\abstraction(x), \abstraction(x'))$.

Model-based approaches promise strategic exploration via planning and fast
transfer to tasks with shared dynamics,
but learning an accurate model over high-dimensional state spaces is challenging.
To avoid compounding errors from function approximation, we propose to construct
an \emph{abstract MDP} over the abstract states defined by $\abstraction$,
which serves as a low-dimensional representation of the world
(we refer to the original MDP, the world, as the concrete MDP).
Concretely, the abstract MDP consists of:

\begin{itemize}
  \item
    \emph{Known set} $\knownset$: a set of abstract states, that can be reached
    (reliably) by executing abstract actions from the start state.

  \item
    \emph{Abstract actions} $\mathcal G$: abstract actions $\go{s}{s'} \in \mathcal{G}$,
    which will be backed by a
    worker policy that output sequences of concrete actions.

  \item
    \emph{Abstract dynamics and rewards} $(P, R)$:
    With probability $P(s' | \go{s}{s'}, s)$,
    executing $\go{s}{s'}$ at abstract state $s$ yields a concrete
    trajectory $(x_0, x_1, \cdots, x_T)$ with $s = \abstraction(x_0)$
    and $s' = \abstraction(x_T)$, and reward $R(s, s')$.
\end{itemize}

Our proposed approach, \ours, learns the abstract MDP as follows:
A manager maintains the known set and estimates of the dynamics $\hat{P}$ and rewards
$\hat{R}$ (\refsec{growing})
with the key property that $\hat{P}$ and $\hat{R}$ are \emph{accurate} on the
$\mathcal{S}$ and $\mathcal{G}$ (the current abstract MDP),
so the manager can plan accurately using the abstract MDP over
significant time horizons without suffering from compounding errors.
Concurrently, we learn a worker policy $\pi(a \mid x, (s, s'))$
which backs the abstract actions of the abstract MDP (\refsec{worker}).
The worker deals with the messy details of the concrete states and actions,
outputting concrete actions $a$ conditioned on the concrete state $x$ to
implement the abstract action $\go{s}{s'}$, allowing the manager to operate
purely on abstract states and abstract actions; the action $\go{s}{s'}$ is
exactly calling the worker at a concrete state $x$ with $\abstraction(x) = s$
on $(s, s')$.

The manager obtains a set of candidate transitions
$\mathcal{C} = \{(s, s'): \text{navigating from } s \text{ to
} s' \text{ is possible}\}$ to grow the abstract MDP via randomized
exploration at the fringes of the known set (\reffig{abstract_mdp}a).
Then, the worker trains on the candidate transitions $(s, s')$ from
$\mathcal{C}$ until it can reliably traverse from $s$ to $s'$, at which point
it adds the action $\go{s}{s'}$ to the action set $\mathcal{G}$
(\reffig{abstract_mdp}b).
The worker maintains an invariant that the action set
$\mathcal{G}$ only contains \emph{reliable} actions: actions $\go{s}{s'}$ such
that $P(s' \mid go(s, s'), s) \geq 1 - \delta$.
This enables the manager to easily revisit the fringes of the known set for
discovering new candidate transitions.
To ensure actions in $\mathcal{G}$ do not later become unreliable from further
updating the worker (forgetting), violating the invariant, the worker learns a
separate skill (neural subpolicy) for each action, reusing skills when
possible.

 \section{Learning the Abstract MDP}\label{sec:growing}

\begin{algorithm}[t]
  \small
  \caption{\textsc{Manager}}\label{alg:manager}
  \begin{algorithmic}[1]
    \WHILE{abstract MDP not fully constructed}
      \STATE Compute the candidate exploration goals $\mathcal{N} \cup
      \{(s, s') \in \mathcal{C}: s \in \mathcal{S}\}$
      \STATE Score all candidates and select highest priority candidate $c$
      \STATE ($c \in C$ is either a transition $(s, s')$ or an abstract state $s$)
      \STATE Compute a plan $\go{s_0}{s_1}, \go{s_1}{s_2}, \cdots, \go{s_{T-1}}{s_T=s}$ with
      model
      \FOR{$t = 1$ to $T$}
        \STATE Call worker to execute $\go{s_{t - 1}}{s_t}$
      \ENDFOR
      \IF{$c$ is a transition $(s, s')$ to learn}
        \STATE reward, success $\gets \textsc{LearnWorker}(s, s')$
        \STATE Update dynamics model $\hat{P}(s, s') \gets$ success rate of past
        $N_{transition}$ attempts
        \STATE Update rewards model $\hat{R}(s, s') \gets \text{reward}$
\ELSE
        \STATE \# $c$ is an abstract state $s$ to explore
        \STATE transitions $\gets \textsc{DiscoverTransitions}()$
        \STATE Add transitions to $C$
      \ENDIF
    \ENDWHILE
  \end{algorithmic}
\end{algorithm}

The manager grows the abstract MDP by adding new actions to $\mathcal{G}$.
It does this in three main steps:
(i) it discovers new candidate transition $(s, s')$ between nearby abstract
states $s$ and $s'$ by randomly exploring around novel states in the known set
$\mathcal{N} \subset \mathcal{S}$ (\refsubsec{discoverer});
(ii) it calls the worker on a candidate transition $(s, s') \in \mathcal{C}$
to learn the abstract action $\go{s}{s'}$: i.e., drive the worker's success
rate $P(s' | \go{s}{s'}, s)$ toward 1;
and (iii) it updates the dynamics model $\hat{P}(s, s')$\footnote{For simplicity, the worker estimates $\hat{P}(s, s')$ instead of $\hat{P}(s' | s, \go{s}{s'})$, effectively treating all failures of $\go{s}{s'}$ navigating
to different abstract state $s'' \neq s'$ equally.},
representing the success rate of abstract action $\go{s}{s'}$, and rewards
model $\hat{R}(s, s')$ (\refalg{manager}).

The manager alternates between (i) and (ii) by maintaining a prioritized list
of \emph{exploration goals} (\refsubsec{priority}), where each goal is either
to discover new transitions around a novel state $s \in \mathcal{N}$ or call
the worker to learn a candidate transition starting in the known set:
$(s, s') \in \mathcal{C}$ with $s \in \mathcal{S}$.
On each episode, the manager selects the highest-priority goal (e.g., the
candidate transition $(s, s')$) (lines 2 - 4),
and navigates to the relevant abstract state (e.g., $s$)
via planning with its dynamics model
(e.g., the plan $\text{go}(s_0, s_1), \cdots,
\text{go}(s_{T - 1}, s_T = s)$) (lines 5 - 8).
Then it executes the selected exploration goal
(e.g., the worker trains on $(s, s')$) (line 10).
Finally, the manager updates its model as follows:
It estimates $\hat{P}(s, s')$ as the success rate of the worker's past
$N_{transition}$ attempts to traverse $(s, s')$ (line 11),
and updates $\hat{R}(s, s')$ as the reward accumulated from the first successful
worker traversal of $(s, s')$ (line 12).

\subsection{Discovering New Transitions}\label{subsec:discoverer}

Unlike the typical model-based RL setting, where the actions are known,
learning the abstract MDP requires simultaneously learning actions for its
action set $\mathcal{G}$.
Thus, the manager must discover candidate transition $(s, s') \in \mathcal{C}$
for the worker to learn as actions $\go{s}{s'}$.

To discover new candidate transitions, the manager revisits a novel abstract
state $s \in \mathcal{N}$ (selected according \refsubsec{priority}),
where novel states are those that have been visited few times:
$\mathcal{N} = \{s \in \mathcal{S}: n(s) < N_{visit}\}$.
Then it follows a simple policy $\pi^d(a_t | x_{0:t}, a_{0:t - 1})$ for $T_d$
timesteps and records the transitions it observes (\refalg{discoverer}).
The policy $\pi^d$ outputs randomized concrete actions $a_t$ conditioned on
the past concrete states $x_{0:t}$ and past concrete actions $a_{0:t - 1}$,
where $\abstraction(x_0) = s$.
During those $T_d$ timesteps, the manager adds the observed transitions
$(\abstraction(x_0), \abstraction(x_1)), \cdots
(\abstraction(x_{T - 1}), \abstraction(x_T))$,
to $\mathcal{C}$.
Additionally, due to imperfections in the abstraction function,
certain candidate transitions may be difficult or impossible for the worker to
learn.
To avoid getting stuck, the manager also adds ``long-distance'' transitions to
$\mathcal{C}$: $(s, s')$ pairs for which the manager did not
directly transition from $s$ to $s'$, but indirectly did so through a sequence
of intermediate states $(s_0 = s, s_1, \cdots, s_T = s')$.
Letting $d(s, s')$ be the length of the shortest such path,
the manager adds all transitions $(s, s')$ with $d(s, s') \leq d_{max}$ to
$\mathcal{C}$.

Navigating to new states often requires repeating the same action multiple
times (e.g., going down the hallway requires going right many times),
so we bias our exploration policy toward repeated actions.
At each timestep, $\pi^d$ uniformly samples a concrete
action and a number between $1$ and $T_{repeat}$, and repeats the action the
sampled number of times.

\subsection{Choosing an Exploration Goal}\label{subsec:priority}

The manager grows the abstract MDP by alternating between executing
exploration goals of two types:
(i) discover new transitions around a novel abstract state $s \in \mathcal{N}$
and (ii) call the worker to learn a candidate transition starting in the known
set: $(s, s') \in \mathcal{C}$ with $s \in \mathcal{S}$ (it must start in the
known set, so that the manager can navigate to the start by planning).
To choose which goal it executes each episode, the manager scores all goals
with a priority function and chooses the goal with the highest priority.
Our theoretical results (\refsec{theory}) hold for any priority function that
eventually chooses all exploration goals.
In practice, we prioritize novel abstract states by their visit count,
prioritizing the abstract states that have been visited fewer times,
yielding the priority function $f_n(s) = -n(s)$.

We prioritize the candidate transitions that are
easy to learn (for fast growth of the abstract MDP) and bottleneck
candidate transitions, transitions that, when learned, enable learning further
transitions.
Candidate transitions that are shorter and are already successfully being
traversed by the worker tend to be easier.
Hence, we estimate how easy a candidate transition is with:
$\lambda_1 n_{succ}(s, s') - n_{fail}(s, s') - d(s, s')^2$,
where $n_{succ}$ is the number of times the worker has successfully traversed
$(s, s')$ and $n_{fail}$ is the number of times the worker has failed in
traversing $(s, s')$.
A candidate $(s, s')$ transition is a bottleneck if learning it is the only
way to reach new abstract states:
i.e., there is another abstract state $s''$ with $(s', s'') \in \mathcal{C}$
and no other candidate transitions end in $s''$.
Altogether, we prioritize a candidate transition as $f_c(s, s') = \lambda_1
n_{succ}(s, s') - n_{fail}(s, s') - d(s, s')^2 + \mathbb{I}_b(s, s') \lambda_2 +
\lambda_3$, where $\mathbb{I}_b(s, s')$ is an indicator that is $1$ if $(s,
s')$ is a bottleneck and $\lambda_3$ is a constant to weight candidate
transitions differently from novel abstract states.
Additionally, to prioritize goals with high reward,
we define $f'_n(s) = f_n(s) + R(s_0, s)$ and $f'_c(s, s') = f_c(s, s') +
R(s_0, s)$,
where $R(s_0, s)$ is the reward received by planning to go from the initial
abstract state $s_0$ to $s$.
Then, the manager alternates switches between $f$ and $f'$.
 \section{Learning the Worker Policy}\label{sec:worker}

\begin{algorithm}[t]
  \small
  \caption{$\textsc{LearnWorker}(s, s', x_0)$}\label{alg:worker}
  \begin{algorithmic}[1]
    \REQUIRE a transition $(s, s')$ to learn, called at
    concrete state $x_0$ with $\abstraction(x_0) = s$
\STATE Set worker horizon $H = d(s, s') \times H_\text{worker}$
    \STATE Choose $a_0 \sim \pi^\text{w}(x_0, (s, s')) =
      \pi_{\mathcal{I}(s, s')}(x_0, s')$
    \FOR{$t = 1$ to $H$}
      \STATE Observe concrete state $x_t$, extrinsic reward $r^E_t$
      \STATE Compute worker intrinsic reward $r^I_t = R_{(s, s')}(x_t | s')$
      \STATE Update worker on $(x_{t - 1}, a_{t - 1}, r^I_t, x_t)$
      \STATE Choose $a_t \sim \pi^\text{w}(x_t, (s, s')) =
        \pi_{\mathcal{I}(s, s')}(x_t, s')$
      \ENDFOR
    \STATE Compute success = $\mathbb{I}[r^I_1 + \dots + r^I_H \geq
    R_\text{hold}]$
    \IF{$\hat{P}(s, s') \geq 1 - \delta$}
      \STATE Freeze worker's skill $\pi_{\mathcal{I}(s, s')}$
      \STATE Add $\go{s}{s'}$ to abstract MDP
    \ENDIF
    \STATE Return extrinsic reward $r^E_1 + \dots + r^E_H$, success
  \end{algorithmic}
\end{algorithm}

The worker handles the messy details of concrete states and
concrete actions so that the manager can plan over the much simpler abstract
state space.
It accomplishes this by learning the subtask of reliably traversing
transitions $(s, s')$ from the manager's candidate transitions $\mathcal{C}$,
thus forming actions $\go{s}{s'}$ in the action set $\mathcal{G}$ of the
abstract MDP.
The worker helps maintain two key properties:
First, to enable the manager to easily visit any abstract state in the known
set,
the worker maintains the invariant that all actions $\go{s}{s'}
\in \mathcal{G}$ are reliable (i.e., $P(s' \mid \go{s}{s'}, s)
\geq 1 - \delta$).
Second, it learns transitions in a way that preserves the Markov property of
the abstract MDP.

While it possible to learn a single policy for all transitions, it is tricky
to ensure that all actions stay reliable, since learning new transitions
can have deleterious effects on previously learned transitions.
Instead, the worker maintains an inventory of \emph{skills} (\refsec{skills}),
where each transition is learned by a single skill, sharing the
same skill amongst many transitions when possible.
The worker uses these skills to form the action set of the abstract MDP
following \refalg{worker}:
When the manager calls the worker on a transition $(s, s')$, the worker
selects the appropriate skill from the skill inventory and begins an episode
of the subtask of traversing $s$ to $s'$ (\refsec{subtask}) (lines 1 - 8).
During the skill episode, the skill receives intrinsic rewards, successfully
completes the subtask if it navigates to $s'$ and meets the worker's
\emph{holding heuristic}, a heuristic for maintaining the Markov property by
ensuring that the worker has not navigated to a concrete state requiring time
information unobserved in the abstract state (line 9).
The worker adds the action $\go{s}{s'}$ to the abstract MDP once it learns to
reliably traverse $(s, s')$ (e.g., $\hat{P}(s, s') \geq 1 - \delta$)
(lines 10 - 13).

\subsection{Skill Repository}\label{sec:skills}
The worker's \emph{skill inventory} $\mathcal{I}$ indexes skills so that
the skill at index $\mathcal{I}(s, s')$ reliably traverses transition $(s, s')$.
Each skill is a goal-conditioned subpolicy $\pi_{\mathcal{I}(s, s')}(a | x,
s')$, which produces concrete actions $a$ conditioned on the current concrete
state $x$ and the goal abstract state $s'$.
When the worker traverses a transition $(s, s')$, it calls on the
corresponding skill until the transition is traversed:
i.e., $\pi^w(a | x, (s, s')) = \pi_{\mathcal{I}(s, s')}(a | x, s')$.

When learning a new transition $(s, s')$, the worker first tries to reuse its already
learned skills from the skill inventory.
For each skill $\pi_i$ in the skill inventory, it measures the success rate of
$\pi_i$ on the new transition $(s, s')$ over $N_{transition}$ attempts.
If the success rate exceeds the reliability threshold $1 - \delta$ for any
skill $\pi_i$,
it updates the skill repository to reuse the skill:
$\mathcal{I}(s, s') \leftarrow \pi_i$.
Otherwise, if no already learned skill can reliably traverse the new
transition, the worker creates a new skill and trains it to navigate the
transition by optimizing intrinsic rewards (\refsec{subtask}).

\subsection{Worker Subtask}\label{sec:subtask}
Given a candidate transition $(s, s')$, the worker's subtask is to navigate from
abstract state $s$ to abstract state $s'$, forming the actions in the
abstract MDP.
Each episode of this subtask consists of $d(s, s') \times H_{worker}$
timesteps (longer transitions need more timesteps to traverse),
where the reward at each timestep is $R_{(s, s')}(x_t) = 1$ if the skill has
successfully reached the end of the transition ($\abstraction(x_t) = s'$) and
0 otherwise.
These episodes additionally terminate if the main episode terminates or if the
manager receives negative environment reward.

When solving these subtasks, the worker must be careful not to violate the
Markov property.
In particular, the concrete state may contain critical
time information unobserved in the abstract state.
For example, consider the task of jumping over a dangerous hole, consisting of
three abstract states: $s_1$ (the cliff before the hole), $s_2$ (the air above
the hole), and $s_3$ (the solid ground on the other side of the hole).
The worker might incorrectly assume that it can reliably traverse from $s_1$
to $s_2$ by simply walking off the cliff.
But adding this as a reliable transition to the abstract MDP causes a problem:
there is now no way to successfully traverse from $s_2$ to $s_3$ due to
missing timing information in the abstract state (i.e., the worker
will soon fall), violating the Markov property.

On navigating a transition $(s, s')$, the worker avoids this problem by
navigating to $s'$ and then checking for
missing timing information with the \emph{holding heuristic}.
The idea is that if critical timing information is missing,
that timing information would eventually cause the abstract state to change
(e.g., in the example, the worker would eventually hit the bottom and die).
Consequently, if the worker can stay in $s'$ for many timesteps, then the
timing must not be important.
This corresponds to only declaring the episode as a success if the worker
accumulates at least $R_{hold}$ reward (equivalent to being in $s'$ for
$R_{hold}$ timesteps).

Any RL algorithm can be used to train the skills to perform this
subtask.
We choose to represent each skill as a Dueling DDQN \citep{van2016deep, wang2016dueling}.
For faster training, the skills use self-imitation
\citep{oh2018self} to more quickly learn from previous successful episodes,
and count-based exploration similar to
\citep{bellemare2016unifying} to more quickly initially discover skill reward.
Since the skill inventory can contain many skills, we save parameters by
occasionally using pixel-blind skills.
\refapp{skill_details} fully describes our skill training and architecture.
 \section{Formal Analysis}\label{sec:theory}

We analyze the \emph{sample complexity} \citep{kakade2003sample} of \ours,
the number of samples required to, with high probability, learn a policy
achieving near-optimal reward.
Standard tabular setting results (e.g., MBIE-EB \citep{strehl2008analysis}) guarantee learning a near-optimal policy, but require a
number of timesteps polynomial in the size of the state space, which is
vacuous in the deep RL setting, with exponentially large state spaces
(e.g., $>10^{100}$ states).

In contrast, assuming that our neural network policy class is rich enough to
represent all necessary skills, with high probability, our approach can learn
a near-optimal policy on a subclass of MDPs in time and space polynomial in
the size of the abstract MDP (details in \refapp{proofs}).
The key intuition is that instead of learning a single long-horizon task,
\ours learns many short-horizon subtasks of navigating from one abstract state
to another.
This is critical, as many deep RL algorithms (e.g. $\epsilon$-greedy) require
a number of samples exponential in the time horizon to solve a task.

 \section{Experiments}\label{sec:experiments}

Following \citep{aytar2018playing}, we empirically evaluate our approach on three
of the most challenging games from the ALE \citep{bellemare2013arcade}:
\mz, \pitfall, and \pe.
We do not evaluate on simpler games (e.g., \textsc{Breakout}), because they are already
solved by prior state-of-the-art methods \citep{hessel2017rainbow} and do not
require sophisticated exploration.
We use the default ALE setup (\refapp{exp_details})
and end the episode when the agent loses a life.
We report rewards from periodic evaluations every 4000 episodes,
where the manager plans for optimal reward in the currently constructed
abstract MDP.
We average our approach over $4$ seeds and report $1$ standard deviation error
bars in the training curves.
Our experiments use the same set of hyperparameters (\refapp{hyperparams}) across all three games,
where the hyperparameters were exclusively and minimally tuned on \mz.

To decouple learning an abstraction from effectively leveraging abstraction,
we use the RAM state (available through the ALE simulator) to extract the
bucketed location of the agent and the agent's inventory for our abstraction
function.
Roughly,
this distinguishes states where the agent is in different locations or has picked
up different items,
but doesn't distinguish states where other details differ
(e.g. monster positions or obstacle configurations).
Notably,
the abstract state function does not specify what each part of the abstract
state means,
and the agent does not know the entire abstract state space beforehand.
We describe the exact abstract states in \refapp{abstract_state_details}.
\refsubsec{adm} also presents a promising direction for automatically learning
the abstraction.

\subsection{Transfer to New Reward Functions}\label{sec:new_rewards}

A key benefit of learning a dynamics model is that it can
facilitate transfer to other reward functions (tasks).
If a new abstract reward function is directly provided,
\ours can perform 0-shot transfer,
by computing a good policy with the learned dynamics and new rewards.
We show here that even if the new reward function is unknown, by leveraging
the learned abstract MDP dynamics model, \ours can quickly transfer
to reward functions that were not seen during training.
It does this by revisiting the abstract MDP's transitions to update the
rewards model with the newly observed reward, and then computing a policy that
achieves high reward under the new reward function via planning.

To evaluate this, we first train \ours on the standard reward function in \mz.
Then, we transfer to 3 new reward functions: (e.g., \textit{kill spider},
which requires the agent to navigate through 5 rooms to pick up and save a
sword for the spider in room 13.)
We compare with \textit{SmartHash} \citep{tang2017exploration}\footnote{
  \textit{SmartHash} was the state-of-the-art exploration
  approach when these experiments were conducted.
},
trained from scratch directly on the new reward functions for 1B (1000M) frames.
Using 1000x fewer frames, \ours successfully transfers: achieving about 3x as
much reward as \textit{SmartHash}.
This transfer can only achieve optimality if the learned abstract MDP
contains all the necessary abstract states in the known set and if the new
rewards only depend on the abstract states (i.e., reward of transitioning from
$x$ to $x'$ is $R(\abstraction(x), \abstraction(x'))$.
However, \ours can still perform well even if these conditions are not met,
highlighting the ability of model-based algorithms
like ours to quickly transfer to new tasks in the same domain.
We describe the details in \refapp{full_new_rewards}.

\subsection{Strategic Exploration}\label{sec:main}

\mz, \pitfall, and \pe have extremely sparse rewards (often hundreds of
timesteps elapse between rewards) and thus require strategic exploration to
solve.
We compare \ours with two main baselines:

\begin{itemize}
  \item
    \textit{AbsHash}: a direct adaptation of \textit{SmartHash}, which uses
    the same RAM information as \ours to perform count-based exploration on
    abstract states (see \citep{tang2017exploration} for details).
  \item
    The non-demonstration state-of-the-art approaches when these experiments
    were conducted:
    in \mz, we compare with \textit{RND} \citep{burda2018exploration};
    in \pitfall, we compare with \textit{SOORL} \citep{keramati2018strategic},
    which is the first non-demonstration approach to achieve positive reward
    on \pitfall, but requires extensive engineering (much stronger than RAM
    state info) to identify and extract objects;
    in \pe, we compare with \textit{DQN-PixelCNN} \citep{ostrovski2017count}.
    Notably, \ours is not directly comparable to these approaches:
    \textit{RND} and \textit{DQN-PixelCNN} do not use RAM information, while
    \textit{SOORL} uses \emph{even stronger} information, but we chose the
    best prior approaches regardless of the prior knowledge they use.
    We also note that methods achieving even higher rewards, e.g.,
    \textit{GoExplore} \citep{ecoffet2019go} and \textit{Atari57}
    \citep{badia2020never} have been released
    since 2018, when these experiments were conducted.
\end{itemize}

\begin{figure*}\centering
    \subfloat[]{{\includegraphics[width=0.27\linewidth]{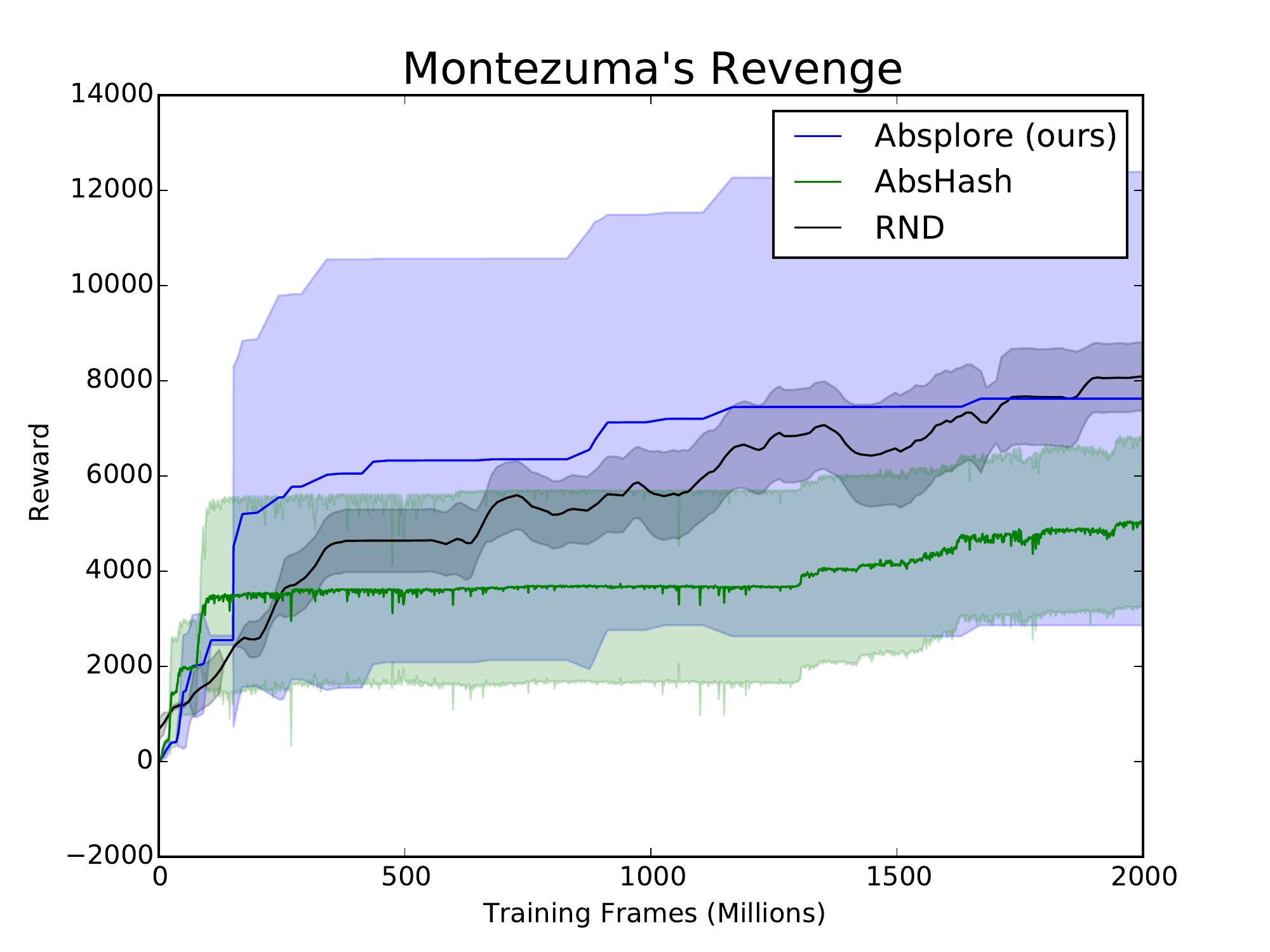} }}\subfloat[]{{\includegraphics[width=0.27\linewidth]{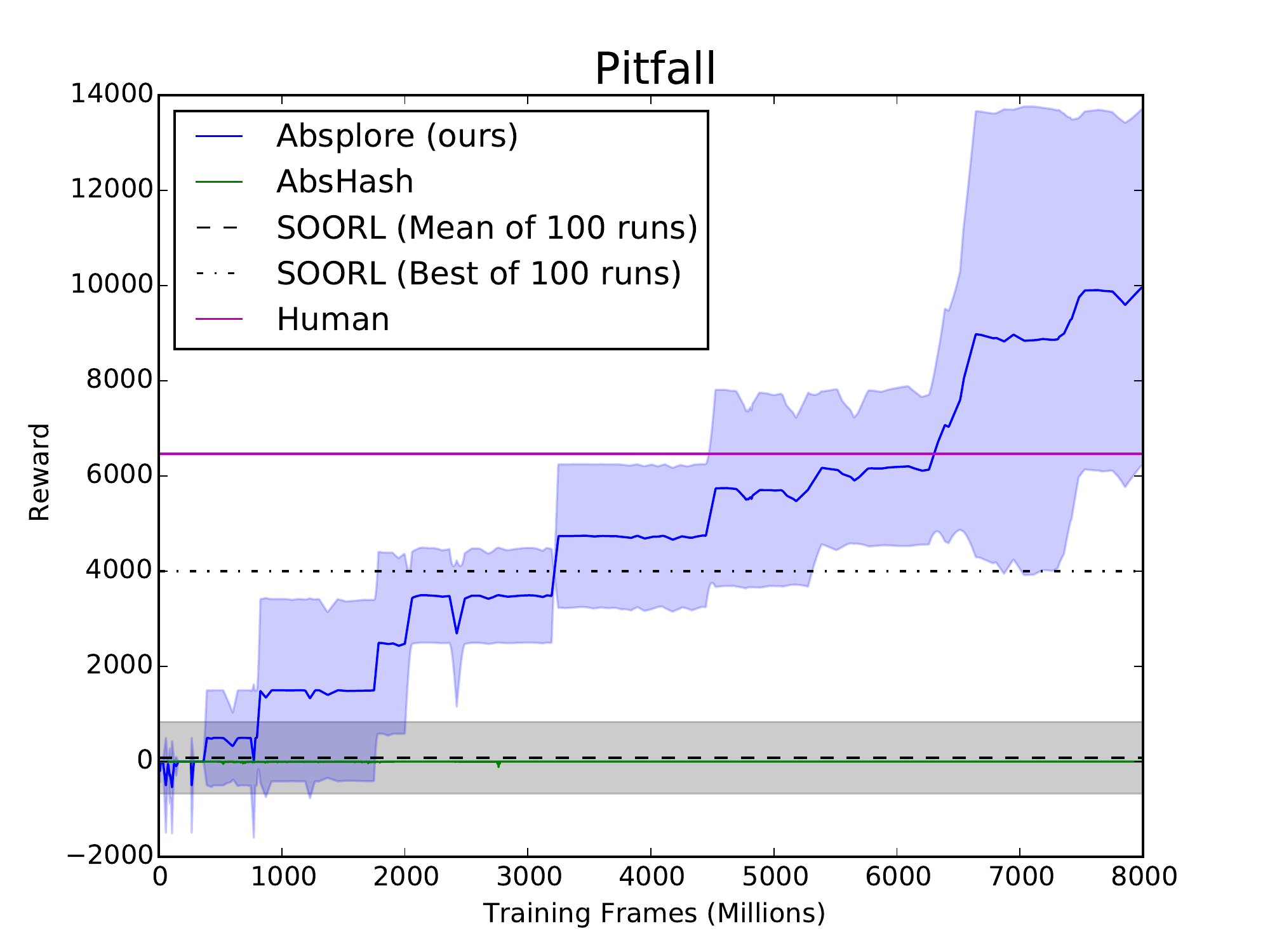} }}\subfloat[]{{\includegraphics[width=0.27\linewidth]{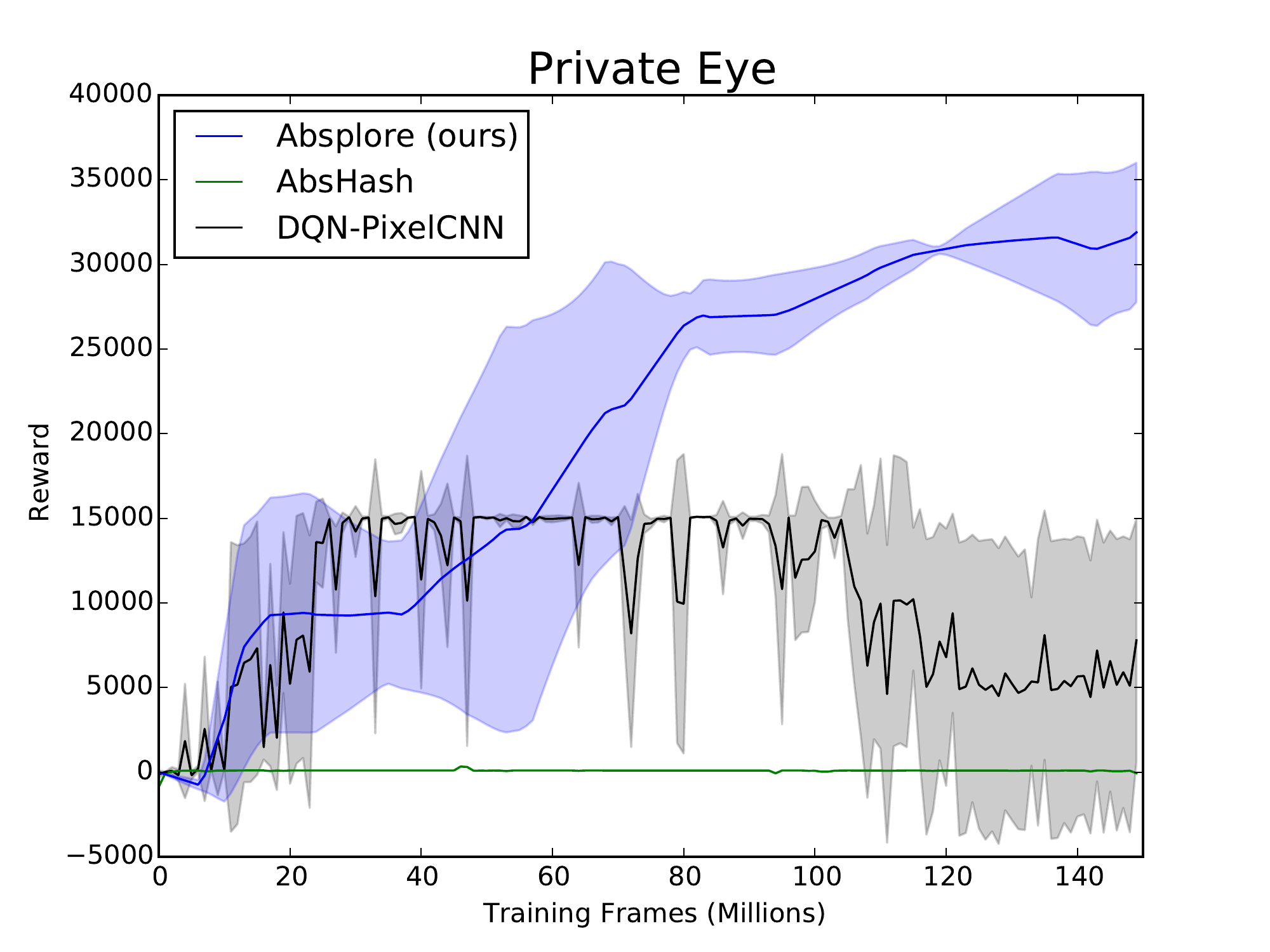} }}\caption{
      Comparison of \ours with non-demonstration state-of-the-art
      approaches in \mz, \pitfall, and \pe.
      \ours achieves superhuman results on \pitfall without demonstrations.
    }\label{fig:main_results}\end{figure*}

\reffig{main_results} shows the main results.
\ours far outperforms \textit{AbsHash}, which suggests that \ours more
effectively leverages abstraction than prior approaches, and that RAM state
does not trivialize the task.
\ours is the (concurrent) first non-demonstration approach to
achieve superhuman performance on \pitfall,
achieving a final average reward of 9959.6 after 8B frames of training,
compared to average human performance: 6464 \citep{pohlen2018observe}.
It achieves more than double the reward of \textit{SOORL},
which achieves a maximum reward of 4000 over 100 seeds and a mean reward of
80.52,
and even significantly outperforms \textit{Ape-X DQfD}
\citep{pohlen2018observe},
which uses high-scoring expert demonstrations during training to achieve a
final mean reward of $3997.5$.
In \mz, after 2B training frames,
\ours achieves comparable but slightly lower average rewards than
\textit{RND},
although \ours continues to improve even late into training and shortly
exceeds the rewards achieved by \textit{RND}.
In \refapp{linear}, we present more results on the ability of \ours
to continue to learn without plateauing.
In \pe,
\ours achieves a mean reward of $32211.5$,
more than double the reward of \textit{DQN-PixelCNN},
which achieves $15806.5$.
Our approach performs even better ($60247$ at 200M frames),
approaching human performance,
when a single hyperparameter is changed (\refapp{private_eye}).

To summarize, \ours successfully uses the abstract MDP for strategic
exploration.
In the three sparsest-reward ALE games, \ours achieves competitive
results, although not directly comparable with \textit{RND}
and \textit{DQN-PixelCNN}, which do not use access to the RAM info used
by \ours.
After we ran our experiments, \textit{GoExplore} \citep{ecoffet2019go}, a
concurrent approach, using similar prior knowledge as RAM info, reported even
higher reward: about 15x more on \mz and 6x more on \pitfall; the main benefit
of \ours over \textit{GoExplore} is its ability to transfer to new tasks and
its ability to learn the abstract MDP.

\subsection{Robustness to Stochasticity}\label{sec:stochasticity}

\begin{figure}[t]
  \centering
  \includegraphics[width=0.66\linewidth]{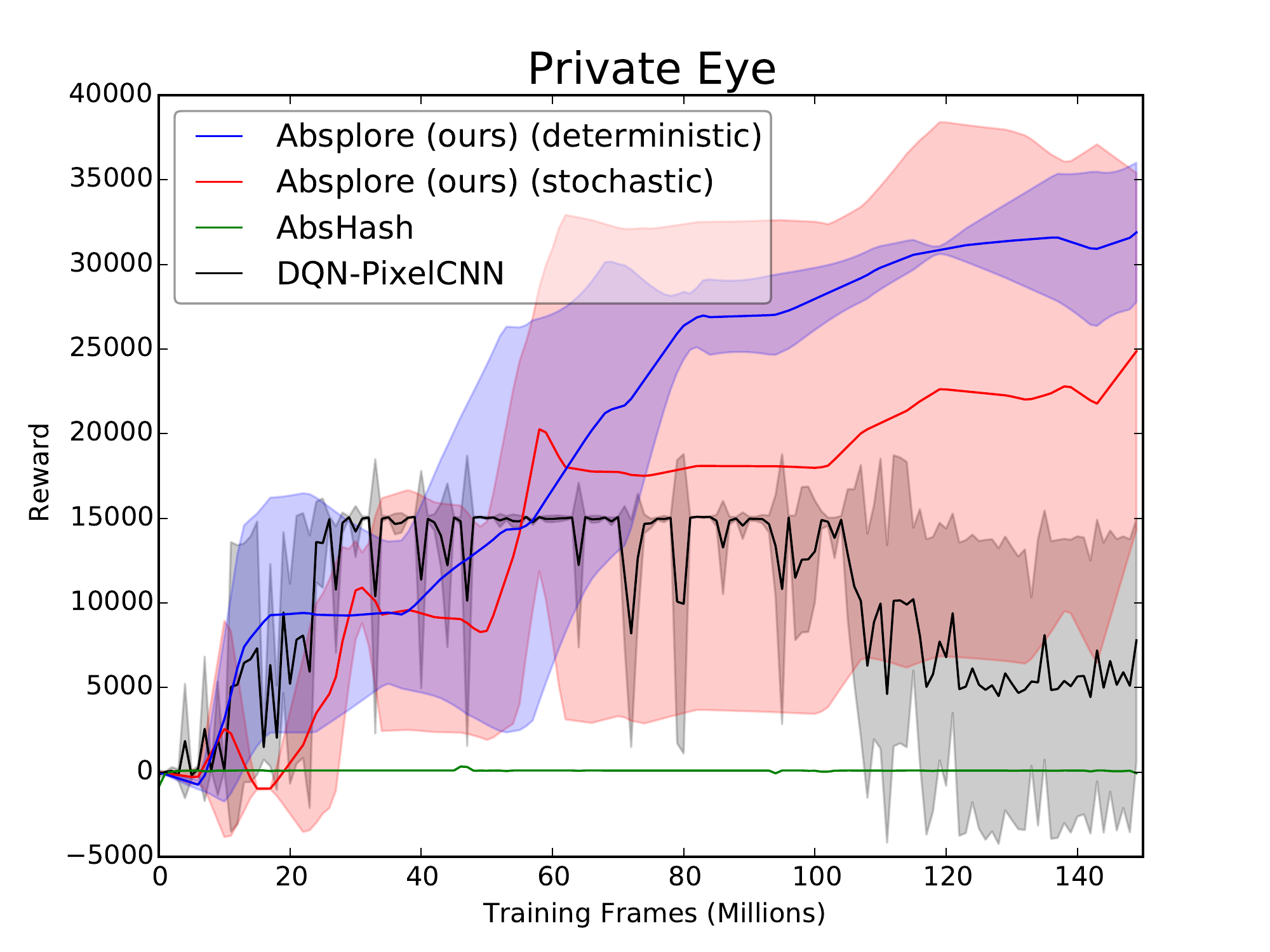}
  \caption{
    \ours continues to outperform the prior state-of-the-art on the
    stochastic version of \pe.
  }
  \label{fig:stochastic}
\end{figure}

We additionally evaluate the performance of our approach on the recommended
\citep{machado2017revisiting}
form of ALE stochasticity (sticky actions 25\% of the time) on \pe
(selected because it requires the fewest frames for training).
\reffig{stochastic} compares the performance of our method on the
stochastic version of \pe with the performance of our method on the
deterministic version of \pe.
Performance degrades slightly on the stochastic version,
because the worker's skills become harder to learn.
However,
both versions outperform the prior state-of-the-art \textit{DQN-PixelCNN},
and the worker is able to successfully abstract away stochasticity
from the manager in the stochastic version of the game
so that the abstract MDP remains near-deterministic.

\subsection{Robustness to Variations in the Abstraction Function}\label{sec:granularity}
An important feature of \ours is its ability to perform well under many
different abstraction functions.
This enables successfully applying it to tasks without searching for the
\emph{perfect} abstraction function.
To study this, we train \ours with multiple variants of our abstraction
function:
Our state abstraction function buckets the agent's $(x, y)$ coordinates.
We obtain variants of it by varying the bucketing size:
increasing the bucketing size results in fewer, coarser abstract states.

\begin{figure}[t]
  \centering
  \includegraphics[width=0.66\linewidth]{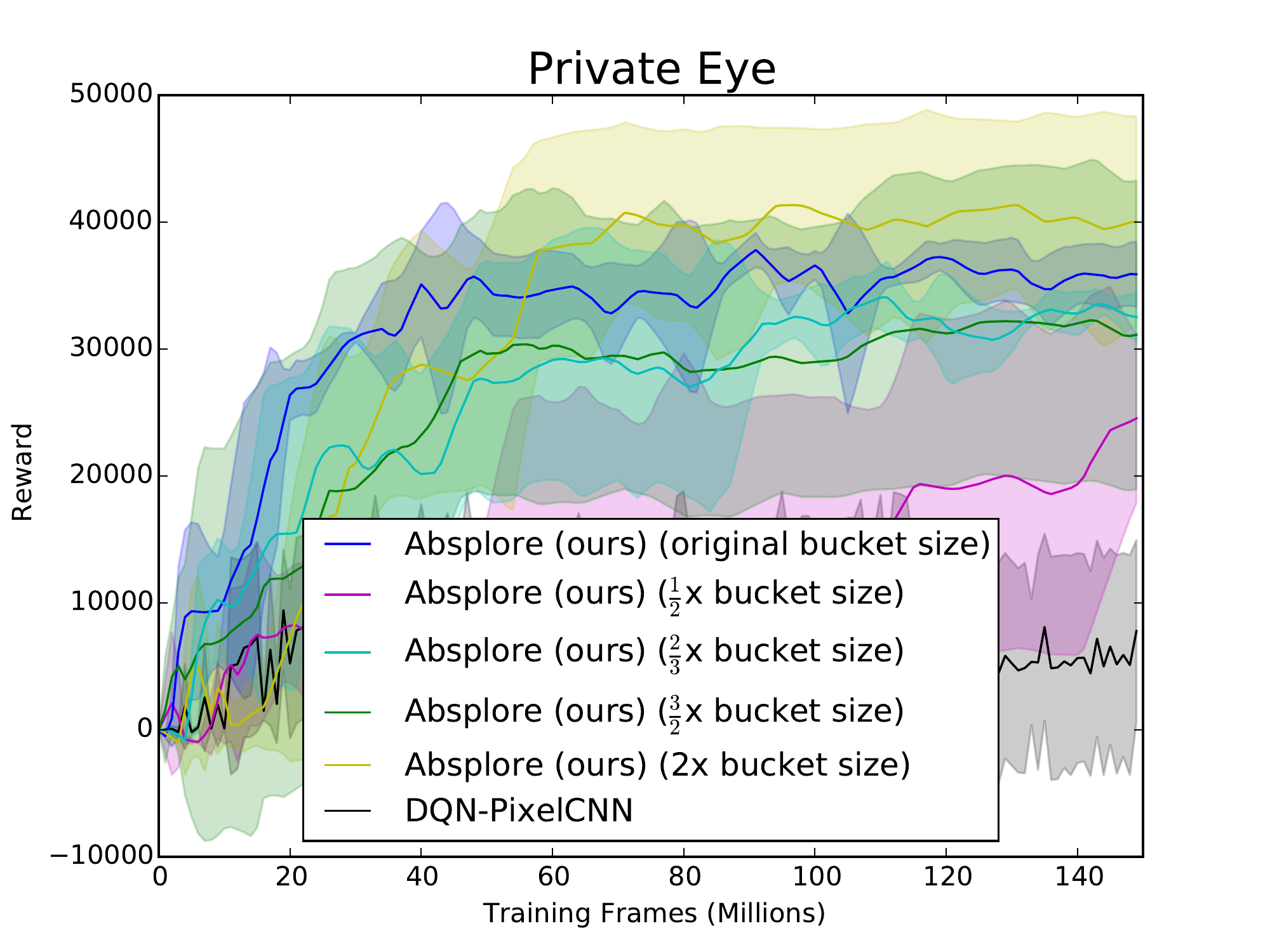}
  \caption{\ours outperforms the prior state-of-the-art on all variants of the
  abstraction function}
  \label{fig:varying}
\end{figure}

We report results in \reffig{varying} on five different bucket sizes obtained
by scaling the original bucket size by $\frac{1}{4}, \frac{4}{9}, 1, \frac{9}{4}$, and $4$.
To adjust for the updated bucket sizes,
we also scale the worker's skill episode horizon $H_{worker}$ by the same
value.
\ours outperforms the prior state-of-the-art approach
\textit{DQN-PixelCNN} across the entire range of bucket sizes,
suggesting that it does not require a highly-tuned state abstraction
function.

\subsection{Automatically Learning the State Abstraction}\label{subsec:adm}

While a drawback of \ours is its use of RAM information for the abstraction
function,
the Attentive Dynamics Model (ADM) \citep{choi2018contingency} presents a
promising avenue for automatically learn the abstraction without RAM information.
Specifically, the key information required from the RAM is the agent's position
(the other parts of our abstraction can be inferred from pixels).
ADM automatically detects the agent's position from pixels by applying a grid
over the screen, predicting the action between consecutive frames in each grid
cell, and learning to attend to the cells that best predict the agent's
action.
The attention naturally focuses on the agent's location,
because the agent's action can only be predicted from the cells containing the
agent.
We compare the abstract states obtained by our RAM state abstraction function
with those obtained by a modified version of ADM on over 100K frames throughout
the rooms of \mz, \pitfall, and \pe and find that they yield similar
abstract states 75\%, 64\%, and 81\% of the time respectively (we defer the
detailed setup of this experiment to \refapp{adm_details}).
While ADM still makes a fair number of errors,
these results are a promising proof-of-concept that the abstraction function
can be automatically learned in future work.
 \section{Related Work}

In theory, model-based RL offers improved exploration \citep{brafman2002r} and faster
transfer to new tasks \citep{laroche2017transfer}.
However, in practice, model-based approaches \citep{guo2014deep},
struggle to match the performance of model-free methods
\citep{hessel2017rainbow} in high-dimensional state spaces, due to errors in
function approximation which compound throughout planning.
More expressive function approximators help for learning accurate models in
relatively low-dimensional (e.g., $< 100$ dimensions) state spaces
\citep{nagabandi2018neural}, but even the best high-dimensional models
\citep{oh2015action, finn2016unsupervised} produce poor longterm
predictions due to compounding errors.
Other works try to robustly plan with imperfect models
\citep{weber2017imagination, abbeel2006using, sutton1990integrated}, but this
reduces sample efficiency.

Our work leverages abstraction \citep{dietterich2000hierarchical, li2006towards}
to learn accurate models.
While many other works \citep{schmidhuber1993planning, singh1995reinforcement,
vezhnevets2017feudal, bacon2017option} similarly leverage abstraction,
they do not learn models, preventing them from planning for strategic
exploration or fast transfer to new reward functions, as we do.
Within the body of hierarchical reinforcement learning (HRL),
two works most relate to ours.
\citet{ecoffet2019go} concurrently developed \textit{GoExplore},
which similarly progressively discovers new abstract states via randomized
exploration on the fringes of the known set.
Unlike \ours, \textit{GoExplore} does not construct a model, which enables
it to be more sample efficient on a single task, but prevents it from
transferring to new tasks.
\citet{roderick2017deep} similarly learn models over abstract states and
skills, but differ in critical design decisions (e.g., sharing parameters
between workers, leading to catastrophic forgetting), which leads to poor
empirical performance.

 \section{Conclusion}
This work presents framework for achieving effective strategic
exploration and fast transfer to shared dynamics tasks via planning.
This is particularly desirable, because many tasks are easily specified with
sparse binary reward functions \citep{nair2017overcoming}, and thus require
strategic exploration.
Also, many real-world families of tasks share the same dynamics: e.g.,
navigating the Internet \citep{shi2017wob, liu2018workflow} or
manipulating objects with a robotic arm.
Whereas prior model-based approaches struggle in high-dimensional state
spaces, due to compound errors from function approximation,
\ours learns an accurate \emph{abstract} model over low-dimensional abstract
states by leveraging the idea of abstraction from hierarchical RL.
This brings the drawback of requiring an abstraction function, but
\refsubsec{adm} presents a promising avenue for future work to automatically
learn the abstraction function.

\paragraph{Reproducibility}

Our code is available at
\url{https://github.com/google-research/google-research/tree/master/strategic_exploration}.
 \subsubsection*{Acknowledgments}

This work was supported by a Google Research Award and a SAIL-Toyota seed
grant.
 
\bibliography{all}

\begin{thebibliography}{54}
\providecommand{\natexlab}[1]{#1}
\providecommand{\url}[1]{\texttt{#1}}
\expandafter\ifx\csname urlstyle\endcsname\relax
  \providecommand{\doi}[1]{doi: #1}\else
  \providecommand{\doi}{doi: \begingroup \urlstyle{rm}\Url}\fi

\bibitem[Abbeel et~al.(2006)Abbeel, Quigley, and Ng]{abbeel2006using}
Abbeel, P., Quigley, M., and Ng, A.~Y.
\newblock Using inaccurate models in reinforcement learning.
\newblock In \emph{International Conference on Machine Learning (ICML)}, pp.\
  1--8, 2006.

\bibitem[Andre \& Russell(2002)Andre and Russell]{andre2002state}
Andre, D. and Russell, S.~J.
\newblock State abstraction for programmable reinforcement learning agents.
\newblock In \emph{Association for the Advancement of Artificial Intelligence
  (AAAI)}, pp.\  119--125, 2002.

\bibitem[Auer et~al.(1995)Auer, Cesa-Bianchi, Freund, and
  Schapire]{auer1995gambling}
Auer, P., Cesa-Bianchi, N., Freund, Y., and Schapire, R.~E.
\newblock Gambling in a rigged casino: The adversarial multi-armed bandit
  problem.
\newblock In \emph{International Conference on Robotics and Automation (ICRA)},
  pp.\  322--322, 1995.

\bibitem[Aytar et~al.(2018)Aytar, Pfaff, Budden, Paine, Wang, and
  de~Freitas]{aytar2018playing}
Aytar, Y., Pfaff, T., Budden, D., Paine, T.~L., Wang, Z., and de~Freitas, N.
\newblock Playing hard exploration games by watching youtube.
\newblock \emph{arXiv preprint arXiv:1805.11592}, 2018.

\bibitem[Bacon et~al.(2017)Bacon, Harb, and Precup]{bacon2017option}
Bacon, P., Harb, J., and Precup, D.
\newblock The option-critic architecture.
\newblock In \emph{Association for the Advancement of Artificial Intelligence
  (AAAI)}, pp.\  1726--1734, 2017.

\bibitem[Badia et~al.(2020)Badia, Sprechmann, Vitvitskyi, Guo, Piot,
  Kapturowski, Tieleman, Arjovsky, Pritzel, Bolt, et~al.]{badia2020never}
Badia, A.~P., Sprechmann, P., Vitvitskyi, A., Guo, D., Piot, B., Kapturowski,
  S., Tieleman, O., Arjovsky, M., Pritzel, A., Bolt, A., et~al.
\newblock Never give up: Learning directed exploration strategies.
\newblock \emph{arXiv preprint arXiv:2002.06038}, 2020.

\bibitem[Bellemare et~al.(2016)Bellemare, Srinivasan, Ostrovski, Schaul,
  Saxton, and Munos]{bellemare2016unifying}
Bellemare, M., Srinivasan, S., Ostrovski, G., Schaul, T., Saxton, D., and
  Munos, R.
\newblock Unifying count-based exploration and intrinsic motivation.
\newblock In \emph{Advances in Neural Information Processing Systems
  (NeurIPS)}, pp.\  1471--1479, 2016.

\bibitem[Bellemare et~al.(2013)Bellemare, Naddaf, Veness, and
  Bowling]{bellemare2013arcade}
Bellemare, M.~G., Naddaf, Y., Veness, J., and Bowling, M.
\newblock The arcade learning environment: An evaluation platform for general
  agents.
\newblock \emph{Journal of Artificial Intelligence Research (JAIR)},
  47:\penalty0 253--279, 2013.

\bibitem[Brafman \& Tennenholtz(2002)Brafman and Tennenholtz]{brafman2002r}
Brafman, R. and Tennenholtz, M.
\newblock {R}-max-a general polynomial time algorithm for near-optimal
  reinforcement learning.
\newblock \emph{Journal of Machine Learning Research}, pp.\  213--231, 2002.

\bibitem[Burda et~al.(2018)Burda, Edwards, Storkey, and
  Klimov]{burda2018exploration}
Burda, Y., Edwards, H., Storkey, A., and Klimov, O.
\newblock Exploration by random network distillation.
\newblock \emph{arXiv preprint arXiv:1810.12894}, 2018.

\bibitem[Chiappa et~al.(2017)Chiappa, Racaniere, Wierstra, and
  Mohamed]{chiappa2017recurrent}
Chiappa, S., Racaniere, S., Wierstra, D., and Mohamed, S.
\newblock Recurrent environment simulators.
\newblock \emph{arXiv preprint arXiv:1704.02254}, 2017.

\bibitem[Choi et~al.(2018)Choi, Guo, Moczulski, Oh, Wu, Norouzi, and
  Lee]{choi2018contingency}
Choi, J., Guo, Y., Moczulski, M., Oh, J., Wu, N., Norouzi, M., and Lee, H.
\newblock Contingency-aware exploration in reinforcement learning.
\newblock \emph{arXiv preprint arXiv:1811.01483}, 2018.

\bibitem[Dietterich(1998)]{dietterich1998maxq}
Dietterich, T.~G.
\newblock The {MAXQ} method for hierarchical reinforcement learning.
\newblock In \emph{International Conference on Machine Learning (ICML)}, 1998.

\bibitem[Dietterich(2000{\natexlab{a}})]{dietterich2000hierarchical}
Dietterich, T.~G.
\newblock Hierarchical reinforcement learning with the {MAXQ} value function
  decomposition.
\newblock \emph{Journal of Artificial Intelligence Research}, pp.\  227--303,
  2000{\natexlab{a}}.

\bibitem[Dietterich(2000{\natexlab{b}})]{dietterich2000state}
Dietterich, T.~G.
\newblock State abstraction in {MAXQ} hierarchical reinforcement learning.
\newblock In \emph{Advances in Neural Information Processing Systems
  (NeurIPS)}, pp.\  994--1000, 2000{\natexlab{b}}.

\bibitem[Ecoffet et~al.(2019)Ecoffet, Huizinga, Lehman, Stanley, and
  Clune]{ecoffet2019go}
Ecoffet, A., Huizinga, J., Lehman, J., Stanley, K.~O., and Clune, J.
\newblock Go-explore: a new approach for hard-exploration problems.
\newblock \emph{arXiv preprint arXiv:1901.10995}, 2019.

\bibitem[Finn et~al.(2016)Finn, Goodfellow, and Levine]{finn2016unsupervised}
Finn, C., Goodfellow, I., and Levine, S.
\newblock Unsupervised learning for physical interaction through video
  prediction.
\newblock In \emph{Advances in neural information processing systems}, pp.\
  64--72, 2016.

\bibitem[Guo et~al.(2014)Guo, Singh, Lee, Lewis, and Wang]{guo2014deep}
Guo, X., Singh, S., Lee, H., Lewis, R.~L., and Wang, X.
\newblock Deep learning for real-time atari game play using offline monte-carlo
  tree search planning.
\newblock In \emph{Advances in neural information processing systems}, pp.\
  3338--3346, 2014.

\bibitem[Henderson et~al.(2017)Henderson, Islam, Bachman, Pineau, Precup, and
  Meger]{henderson2017deep}
Henderson, P., Islam, R., Bachman, P., Pineau, J., Precup, D., and Meger, D.
\newblock Deep reinforcement learning that matters.
\newblock \emph{arXiv preprint arXiv:1709.06560}, 2017.

\bibitem[Hessel et~al.(2017)Hessel, Modayil, Hasselt, Schaul, Ostrovski,
  Dabney, Horgan, Piot, Azar, and Silver]{hessel2017rainbow}
Hessel, M., Modayil, J., Hasselt, H.~V., Schaul, T., Ostrovski, G., Dabney, W.,
  Horgan, D., Piot, B., Azar, M., and Silver, D.
\newblock Rainbow: Combining improvements in deep reinforcement learning.
\newblock \emph{arXiv preprint arXiv:1710.02298}, 2017.

\bibitem[Hester et~al.(2018)Hester, Vecerik, Pietquin, Lanctot, Schaul, Piot,
  Sendonaris, Dulac{-}Arnold, Osband, Agapiou, Leibo, and
  Gruslys]{hester2018deep}
Hester, T., Vecerik, M., Pietquin, O., Lanctot, M., Schaul, T., Piot, B.,
  Sendonaris, A., Dulac{-}Arnold, G., Osband, I., Agapiou, J., Leibo, J.~Z.,
  and Gruslys, A.
\newblock Deep {Q}-learning from demonstrations.
\newblock In \emph{Association for the Advancement of Artificial Intelligence
  (AAAI)}, 2018.

\bibitem[Kakade et~al.(2003)]{kakade2003sample}
Kakade, S.~M. et~al.
\newblock \emph{On the sample complexity of reinforcement learning}.
\newblock PhD thesis, University of London, 2003.

\bibitem[Kearns \& Singh(2002)Kearns and Singh]{kearns2002near}
Kearns, M. and Singh, S.
\newblock Near-optimal reinforcement learning in polynomial time.
\newblock \emph{Machine Learning}, 49\penalty0 (2):\penalty0 209--232, 2002.

\bibitem[Keramati et~al.(2018)Keramati, Whang, Cho, and
  Brunskill]{keramati2018strategic}
Keramati, R., Whang, J., Cho, P., and Brunskill, E.
\newblock Strategic object oriented reinforcement learning.
\newblock \emph{arXiv preprint arXiv:1806.00175}, 2018.

\bibitem[Kingma \& Ba(2014)Kingma and Ba]{kingma2014adam}
Kingma, D. and Ba, J.
\newblock Adam: A method for stochastic optimization.
\newblock \emph{arXiv preprint arXiv:1412.6980}, 2014.

\bibitem[Kirkpatrick et~al.(2017)Kirkpatrick, Pascanu, Rabinowitz, Veness,
  Desjardins, Rusu, Milan, Quan, Ramalho, Grabska-Barwinska,
  et~al.]{kirkpatrick2017overcoming}
Kirkpatrick, J., Pascanu, R., Rabinowitz, N., Veness, J., Desjardins, G., Rusu,
  A.~A., Milan, K., Quan, J., Ramalho, T., Grabska-Barwinska, A., et~al.
\newblock Overcoming catastrophic forgetting in neural networks.
\newblock \emph{Proceedings of the national academy of sciences}, 2017.

\bibitem[Laroche \& Barlier(2017)Laroche and Barlier]{laroche2017transfer}
Laroche, R. and Barlier, M.
\newblock Transfer reinforcement learning with shared dynamics.
\newblock In \emph{Association for the Advancement of Artificial Intelligence
  (AAAI)}, pp.\  2147--2153, 2017.

\bibitem[Li et~al.(2006)Li, Walsh, and Littman]{li2006towards}
Li, L., Walsh, T.~J., and Littman, M.~L.
\newblock Towards a unified theory of state abstraction for mdps.
\newblock In \emph{International Symposium on Artificial Intelligence and
  Mathematics (ISAIM)}, 2006.

\bibitem[Liu et~al.(2018)Liu, Guu, Pasupat, Shi, and Liang]{liu2018workflow}
Liu, E.~Z., Guu, K., Pasupat, P., Shi, T., and Liang, P.
\newblock Reinforcement learning on web interfaces using workflow-guided
  exploration.
\newblock In \emph{International Conference on Learning Representations
  (ICLR)}, 2018.

\bibitem[Machado et~al.(2017)Machado, Bellemare, Talvitie, Veness, Hausknecht,
  and Bowling]{machado2017revisiting}
Machado, M.~C., Bellemare, M.~G., Talvitie, E., Veness, J., Hausknecht, M., and
  Bowling, M.
\newblock Revisiting the arcade learning environment: Evaluation protocols and
  open problems for general agents.
\newblock \emph{arXiv preprint arXiv:1709.06009}, 2017.

\bibitem[Martins \& Astudillo(2016)Martins and Astudillo]{martins2016softmax}
Martins, A. and Astudillo, R.
\newblock From softmax to sparsemax: A sparse model of attention and
  multi-label classification.
\newblock In \emph{International Conference on Machine Learning (ICML)}, pp.\
  1614--1623, 2016.

\bibitem[Mnih et~al.(2015)Mnih, Kavukcuoglu, Silver, Rusu, Veness, Bellemare,
  Graves, Riedmiller, Fidjeland, Ostrovski, et~al.]{mnih2015human}
Mnih, V., Kavukcuoglu, K., Silver, D., Rusu, A.~A., Veness, J., Bellemare,
  M.~G., Graves, A., Riedmiller, M., Fidjeland, A.~K., Ostrovski, G., et~al.
\newblock Human-level control through deep reinforcement learning.
\newblock \emph{Nature}, 518\penalty0 (7540):\penalty0 529--533, 2015.

\bibitem[Nagabandi et~al.(2018)Nagabandi, Kahn, Fearing, and
  Levine]{nagabandi2018neural}
Nagabandi, A., Kahn, G., Fearing, R.~S., and Levine, S.
\newblock Neural network dynamics for model-based deep reinforcement learning
  with model-free fine-tuning.
\newblock In \emph{International Conference on Robotics and Automation (ICRA)},
  pp.\  7559--7566, 2018.

\bibitem[Nair et~al.(2017)Nair, McGrew, Andrychowicz, Zaremba, and
  Abbeel]{nair2017overcoming}
Nair, A., McGrew, B., Andrychowicz, M., Zaremba, W., and Abbeel, P.
\newblock Overcoming exploration in reinforcement learning with demonstrations.
\newblock \emph{arXiv preprint arXiv:1709.10089}, 2017.

\bibitem[Nair \& Hinton(2010)Nair and Hinton]{nair2010rectified}
Nair, V. and Hinton, G.~E.
\newblock Rectified linear units improve restricted {boltzmann} machines.
\newblock In \emph{International Conference on Machine Learning (ICML)}, pp.\
  807--814, 2010.

\bibitem[Oh et~al.(2015)Oh, Guo, Lee, Lewis, and Singh]{oh2015action}
Oh, J., Guo, X., Lee, H., Lewis, R.~L., and Singh, S.
\newblock Action-conditional video prediction using deep networks in atari
  games.
\newblock In \emph{Advances in neural information processing systems}, pp.\
  2863--2871, 2015.

\bibitem[Oh et~al.(2018)Oh, Guo, Singh, and Lee]{oh2018self}
Oh, J., Guo, Y., Singh, S., and Lee, H.
\newblock Self-imitation learning.
\newblock \emph{arXiv preprint arXiv::1806.05635}, 2018.

\bibitem[Ostrovski et~al.(2017)Ostrovski, Bellemare, Oord, and
  Munos]{ostrovski2017count}
Ostrovski, G., Bellemare, M.~G., Oord, A., and Munos, R.
\newblock Count-based exploration with neural density models.
\newblock In \emph{International Conference on Machine Learning (ICML)}, pp.\
  2721--2730, 2017.

\bibitem[Pohlen et~al.(2018)Pohlen, Piot, Hester, Azar, Horgan, Budden,
  Barth-Maron, van Hasselt, Quan, Ve{\v{c}}er{'\i}k, et~al.]{pohlen2018observe}
Pohlen, T., Piot, B., Hester, T., Azar, M.~G., Horgan, D., Budden, D.,
  Barth-Maron, G., van Hasselt, H., Quan, J., Ve{\v{c}}er{'\i}k, M., et~al.
\newblock Observe and look further: Achieving consistent performance on
  {ATARI}.
\newblock \emph{arXiv preprint arXiv:1805.11593}, 2018.

\bibitem[Roderick et~al.(2017)Roderick, Grimm, and Tellex]{roderick2017deep}
Roderick, M., Grimm, C., and Tellex, S.
\newblock Deep abstract {Q}-networks.
\newblock \emph{arXiv preprint arXiv:1710.00459}, 2017.

\bibitem[Schmidhuber(1993)]{schmidhuber1993planning}
Schmidhuber, J.
\newblock Planning simple trajectories using neural subgoal generators.
\newblock In \emph{From Animals to Animats 2: Proceedings of the Second
  International Conference on Simulation of Adaptive Behavior}, volume~2, 1993.

\bibitem[Shi et~al.(2017)Shi, Karpathy, Fan, Hernandez, and Liang]{shi2017wob}
Shi, T., Karpathy, A., Fan, L., Hernandez, J., and Liang, P.
\newblock World of bits: An open-domain platform for web-based agents.
\newblock In \emph{International Conference on Machine Learning (ICML)}, 2017.

\bibitem[Singh et~al.(1995)Singh, Jaakkola, and Jordan]{singh1995reinforcement}
Singh, S.~P., Jaakkola, T., and Jordan, M.
\newblock Reinforcement learning with soft state aggregation.
\newblock \emph{Advances in neural information processing systems}, pp.\
  361--368, 1995.

\bibitem[Strehl \& Littman(2008)Strehl and Littman]{strehl2008analysis}
Strehl, A.~L. and Littman, M.~L.
\newblock An analysis of model-based interval estimation for markov decision
  processes.
\newblock \emph{Journal of Computer and System Sciences}, 74\penalty0
  (8):\penalty0 1309--1331, 2008.

\bibitem[Sutton(1990)]{sutton1990integrated}
Sutton, R.~S.
\newblock Integrated architectures for learning, planning, and reacting based
  on approximating dynamic programming.
\newblock \emph{Machine Learning Proceedings}, pp.\  216--224, 1990.

\bibitem[Sutton et~al.(1999)Sutton, Precup, and Singh]{sutton1999between}
Sutton, R.~S., Precup, D., and Singh, S.
\newblock Between mdps and semi-mdps: A framework for temporal abstraction in
  reinforcement learning.
\newblock \emph{Articial intelligence}, 112:\penalty0 181--211, 1999.

\bibitem[Talvitie(2014)]{talvitie2014model}
Talvitie, E.
\newblock Model regularization for stable sample rollouts.
\newblock In \emph{Uncertainty in Artificial Intelligence (UAI)}, pp.\
  780--789, 2014.

\bibitem[Talvitie(2015)]{talvitie2015agnostic}
Talvitie, E.
\newblock Agnostic system identification for monte carlo planning.
\newblock In \emph{Association for the Advancement of Artificial Intelligence
  (AAAI)}, pp.\  2986--2992, 2015.

\bibitem[Tang et~al.(2017)Tang, Houthooft, Foote, Stooke, Chen, Duan, Schulman,
  DeTurck, and Abbeel]{tang2017exploration}
Tang, H., Houthooft, R., Foote, D., Stooke, A., Chen, X., Duan, Y., Schulman,
  J., DeTurck, F., and Abbeel, P.
\newblock \#exploration: A study of count-based exploration for deep
  reinforcement learning.
\newblock In \emph{Advances in Neural Information Processing Systems
  (NeurIPS)}, pp.\  2753--2762, 2017.

\bibitem[van Hasselt et~al.(2016)van Hasselt, Guez, and Silver]{van2016deep}
van Hasselt, H., Guez, A., and Silver, D.
\newblock Deep reinforcement learning with double {Q}-learning.
\newblock In \emph{Association for the Advancement of Artificial Intelligence
  (AAAI)}, volume~16, pp.\  2094--2100, 2016.

\bibitem[Vezhnevets et~al.(2017)Vezhnevets, Osindero, Schaul, Heess, Jaderberg,
  Silver, and Kavukcuoglu]{vezhnevets2017feudal}
Vezhnevets, A.~S., Osindero, S., Schaul, T., Heess, N., Jaderberg, M., Silver,
  D., and Kavukcuoglu, K.
\newblock Feudal networks for hierarchical reinforcement learning.
\newblock \emph{arXiv preprint arXiv:1703.01161}, 2017.

\bibitem[Wang et~al.(2016)Wang, Schaul, Hessel, Hasselt, Lanctot, and
  Freitas]{wang2016dueling}
Wang, Z., Schaul, T., Hessel, M., Hasselt, H.~V., Lanctot, M., and Freitas,
  N.~D.
\newblock Dueling network architectures for deep reinforcement learning.
\newblock In \emph{International Conference on Machine Learning (ICML)}, 2016.

\bibitem[Watkins(1989)]{watkins1989learning}
Watkins, C.
\newblock Learning from delayed rewards.
\newblock \emph{King's College, Cambridge}, 1989.

\bibitem[Weber et~al.(2017)Weber, Racani{\`e}re, Reichert, Buesing, Guez,
  Rezende, Badia, Vinyals, Heess, Li, et~al.]{weber2017imagination}
Weber, T., Racani{\`e}re, S., Reichert, D.~P., Buesing, L., Guez, A., Rezende,
  D.~J., Badia, A.~P., Vinyals, O., Heess, N., Li, Y., et~al.
\newblock Imagination-augmented agents for deep reinforcement learning.
\newblock \emph{arXiv preprint arXiv:1707.06203}, 2017.

\end{thebibliography}
\bibliographystyle{icml2019}

\clearpage
\appendix

\section{Experiment Details}\label{sec:exp_details}
Following \citet{mnih2015human},
the pixel concrete states are downsampled and cropped to $84$ by $84$
and then are converted to grayscale.
To capture velocity information,
the worker receives as input the past four frames stacked together.
Every action is repeated $4$ times.

In addition,
\mz and \pitfall are deterministic by default.
As a result,
the manager deterministically navigates to the fringes of the known set
by calling on the worker's deterministic, saved skills.
To minimize wallclock training time,
we save the states at the fringes of the known set and enable the worker to
teleport to those states,
instead of repeatedly re-simulating the entire trajectory.
When the worker teleports,
we count all the frames it would have had to simulate as part of the training
frames.
Importantly,
this only affects wallclock time,
and does not benefit or change the agent in any way.
Notably,
this does not apply to \pe,
where the initial state is stochastically chosen from two similar possible
states.

\subsection{Hyperparameters}\label{sec:hyperparams}
All of our hyperparameters are only tuned on \mz.
Our skills are trained with the Adam optimizer \citep{kingma2014adam} with
the default hyperparameters.
\reftab{hyperparameters} describes all hyperparameters and the values
used during experiments (bolded),
as well as other values that we tuned over (non-bolded).
Most of our hyperparameters were selected once and never tuned.

\begin{table*}
\begin{center}
\begin{tabular}{ |c|c| }
 \hline
 Hyperparameter & Value \\
 \hline
 Success weight $\lambda_1$ & (\textbf{1} (stochastic), 10, \textbf{100}
 (deterministic)) \\
 New transition exploration goal weight $\lambda_2$ & \textbf{5000} \\
 Abstract state exploration goal weight $\lambda_3$ & \textbf{-2000} \\
 \hline
 Discovery exploration horizon $T_d$ & \textbf{50} \\
 Discovery visit threshold $N_{visit}$ & \textbf{500} \\
 Discovery repeat action range & 1 to 10, \textbf{1 to 20}, 1 to 30 \\
 \hline
 Worker horizon $H_{worker}$ & (10, 15, 20, \textbf{30}, 45) \\
 Skill failure tolerance $\delta$ & (\textbf{0.05}, 0.1) \\
 Skill holding heuristic $R_{hold}$ & \textbf{4} \\
 Maximum transition distance $d_{max}$ & \textbf{15} \\
 Dynamics sliding window size $N_{transition}$ & \textbf{100} \\
 \hline
 Adam learning rate & \textbf{0.001} \\
 Max buffer size of each skill & \textbf{5000} \\
 Skill DQN target sync frequency & \textbf{75} \\
 Skill batch size & \textbf{32} \\
 Skill minimum buffer size & \textbf{50} \\
 Gradient norm clipping & \textbf{3.0} \\
 Count-based weight $\beta$ & \textbf{0.63} \\
 Margin weight $\lambda$ & \textbf{0.5} \\

 \hline
\end{tabular}
\caption{
  Table of all hyperparameters and the values used in the experiments.
}
\label{tab:hyperparameters}
\end{center}
\end{table*}

\subsection{State Abstraction Function}\label{sec:abstract_state_details}
In \mz,
each abstract state is a
(bucketed agent x-coordinate, bucketed agent y-coordinate, agent room number,
agent inventory, current room objects, agent inventory history) tuple.
These are given by the RAM state at indices 42 (bucketed by 20), 43 (bucketed by 20),
3, 65, and 66 respectively.
The agent inventory history is a counter of the number of times the current
room objects change
(the room objects change when the agent picks up an object).

In \pitfall,
each abstract state is a
(bucketed agent x-coordinate, bucketed agent y-coordinate, agent room number,
items that the agent has picked up) tuple.
These are given by the RAM state at indices 97 (bucketed by 20), 105 (bucketed
by 20), 1, and 113 respectively.

In \pe,
each abstract state is a
(bucketed agent x-coordinate, bucketed agent y-coordinate, agent room number,
agent inventory, agent inventory history, tasks completed by the agent) tuple.
These are given by the RAM state at indices 63 (bucketed by 40),
86 (bucketed by 20),
92, 60, 72, and 93 respectively.

\subsection{Skill Training and Architecture}\label{sec:skill_details}
\paragraph{Architecture.}
Our skills are represented as Dueling DDQNs
\citep{van2016deep, wang2016dueling},
which produce the state-action value
$Q_{(s, s')}(x, a) = A_{(s, s')}(x, a) + V_{(s, s')}(x)$,
where $A_{(s, s')}(x, a)$ is the advantage and
$V_{(s, s')}(x)$ is the state-value function.
The skills recover a policy $\pi_{K(s, s')}(a | x, (s, s'))$ by greedily
selecting the action with the highest Q-value at each concrete state $x$.

The skill uses the standard architecture \citep{mnih2015human} to represent
$A_{(s, s')}(x, a)$ and $V_{(s, s')}(x)$ with a small modification to also
condition on the transition $(s, s')$.
First,
after applying the standard ALE pre-processing,
the skill computes the pixel embedding $e_x$ of the pixel state $x$
by applying three square convolutional layers with (filters, size, stride)
equal to $(32, 8, 4), (64, 4, 2)$, and $(64, 4, 2)$ respectively with
rectifier non-linearities \citep{nair2010rectified},
and applying a final rectified linear layer with output size $512$.
Next,
the skill computes the transition embedding $e_{(s, s'})$ by
concatenating $[e_r; e_{diff}]$ and applying a final rectified linear layer
with output size $64$,
where:

\begin{itemize}
\item
  $e_r$ is computed as the cumulative reward received by the skill during the
  skill episode,
  represented as one-hot,
  and passed through a single rectified linear layer of output size $32$.

\item
  $e_{diff}$ is computed as $s' - s$ passed through a single rectified linear
  layer of output size $96$.
\end{itemize}

Finally,
$e_x$ and $e_{(s, s')}$ are concatenated and passed through a final linear layer to obtain
$A_{(s, s')}(x, a)$ and $V_{(s, s')}(x)$.

To prevent the skill from changing rapidly as it begins to converge on the
optimal policy,
we keep a sliding window estimate of its success rate $p_{success}$.
At each timestep,
with probability $1 - p_{success}$,
we sample a batch of $(x, a, r, x')$ tuples for transition $(s, s')$ from the
replay buffer and update the policy according the DDQN loss function:
$\mathcal{L} = ||Q_{(s, s')}(x, a) - \text{target}||_2^2$,
where
$\text{target} = (r + Q^\text{target}(x', \argmax_{a' \in \mathcal{A}}{Q_{(s, s')}(x', a'))})$.
Additionally,
since the rewards are intrinsically given,
the optimal Q-value is known to be between $0$ and $R_{hold}$.
We increase stability by clipping target between these values.

\paragraph{Pixel blindness.}
In addition,
some skills are easy to learn (e.g. move a few steps to the left)
and don't require pixel inputs to learn at all.
To prevent the skills from unnecessarily using millions of
parameters for these easy skills,
the worker first attempts to learn \textit{pixel-blind} skills
for simple transitions $(s, s')$ with $d(s, s') = 1$
(i.e. $(s, s')$ was directly observed by the manager).
The pixel-blind skills only compute $e_{(s, s')}$ and pass this
through a final layer to compute the advantage and value functions
(they do not compute or concatenate with $e_x$).
If the worker fails to learn a pixel-blind skill,
(e.g. if the skill actually requires pixel inputs,
such as jumping over a monster)
it will later try to learn a pixel-aware skill instead.

\paragraph{Epsilon schedule.}
The skills use epsilon-greedy exploration,
where at each timestep,
with probability $\epsilon$,
a random action is selected instead of the one produced by the skill's policy
\citep{watkins1989learning}.
Once a skill becomes frozen,
$\epsilon$ is permanently set to $0$.

\begin{figure}
    \centering
    \includegraphics[width=0.6\linewidth]{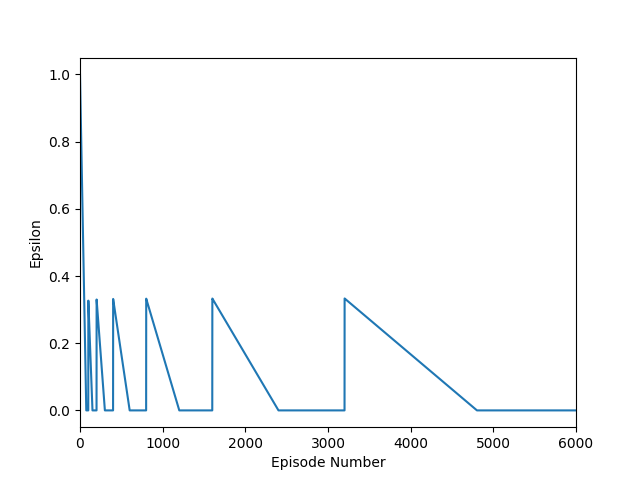}
    \caption{The saw-tooth epsilon schedule used by our skills}
    \label{fig:epsilon}
\end{figure}

The number of episodes required to learn each skill is not known in advance,
since some skills require many episodes to learn
(e.g. traversing a difficult obstacle),
while other skills learn in few episodes (e.g. moving a little to the left).
Because of this,
using an epsilon schedule that decays over a fixed number of episodes,
which is typical for many RL algorithms,
is insufficient.
If epsilon is decayed over too many episodes,
the simple skills waste valuable training time making exploratory actions,
even though they've already learned near-optimal behavior.
In contrast,
if epsilon is decayed over too few episodes,
the most difficult skills may never observe reward,
and may consequently fail to learn.
To address this,
we draw motivation from the doubling trick in online learning \cite{auer1995gambling}
to create an epsilon schedule,
which accomodates skills requiring varying number of episodes to learn.
Instead of choosing a fixed horizon,
we decay epsilon over horizons of exponentially increasing length,
summarized in \reffig{epsilon}.
This enables skills that learn quickly to achieve low values of epsilon early
on in training,
while skills that learn slowly will later explore with high values of epsilon
over many episodes.

\paragraph{Count-based exploration.}
Our skill additionally use count-based exploration similar to
\citet{tang2017exploration, bellemare2016unifying} to learn more quickly.
Each skill maintains a count of the number of times $\text{visit}(s)$ it has
visited each abstract state $s$.
Then,
the skill provides itself with additional intrinsic reward to motivate itself
to visit novel states,
equal to $\frac{\beta}{\sqrt{\text{visit}(s)}}$ each time it visits abstract
state $s$.
We choose $\beta = 0.63$,
an intrinsic reward of approximately
$\frac{2}{\sqrt{10 \times \text{visit}(s)}}$.

\paragraph{Self-imitation.}
When learning to traverse difficult obstacles
(e.g. jumping over a disappearing floor),
the skill may observe a few successes long before successfully learning a
policy to reliably traverse the difficult obstacle.
We use a variant of the self-imitation described in \citep{oh2018self} to decrease this time.
Whenever a skill successfully traverses a transition,
it adds the entire successful trajectory to a separate replay buffer and
performs imitation learning on the successful trajectories.
These successful trajectories are actually optimal skill trajectories
because the skill episode uses undiscounted reward,
so all successful trajectories are equally optimal.
To update on these skills,
the skill periodically samples from this replay buffer and updates on an
imitation loss function
$\mathcal{L}_{imitation}(\theta) = \mathcal{L}_1(\theta) +
\mathcal{L}_2(\theta)$,
where $\theta$ is the skill's parameters,
and $\mathcal{L}_1$ and $\mathcal{L}_2$ are defined as below:

\begin{itemize}
  \item Let $G_t = \sum_{i = t}^T r_t$ be the reward to-go for a successful
    trajectory $(x_0, a_0, r_0), \cdots, (x_T, a_T, r_T)$.
    $\mathcal{L}_1$ directly regresses $Q_{(s, s')}(x, a)$ on the reward
    to-go of the successful trajectory,
    because $G_t$ is actually the optimal Q-value on successful trajectories
    (all successful trajectories are equally optimal): i.e.,
    $\mathcal{L}_1 = ||G_t - Q_{(s, s')}(x_t, a_t)||_2$.

  \item
    We use the margin-loss from \citet{hester2018deep} for
    $\mathcal{L}_2$.
    When sampling a transition $(x, a_E, r, x')$,
    $\mathcal{L}_2 = \max_{a \in \mathcal{A}}[Q_{(s, s')}(x, a) + \lambda \1[{a
    = a_E}]] - Q_{(s, s')}(s, a_E)$.
    Intuitively,
    $\mathcal{L}_2$ encourages the skill to replay the actions that led to
    successful trajectories over other actions.
    We use $\lambda = 0.5$,
    which was chosen with no hyperparameter tuning.
\end{itemize}

\section{Final Performance}\label{sec:exp_perf}
Table \ref{tab:perf} compares the final performance of our method with the
previous state-of-the-art after training.
\begin{table*}
\small{
\begin{center}
    \begin{tabular}{ | c | c | c |}
    \hline
    Envinronment & \ours & State-of-the-art\\ \hline
    \mz (2B Frames)& 7625 & \textbf{8152} (\textit{RND}
    \cite{burda2018exploration}) \\ \hline
    \pitfall (8B Frames)& \textbf{9959.6} & 80.52 (\textit{SOORL}
    \cite{keramati2018strategic}) \\\hline
    \pe (150M Frames)& \textbf{32211.5} & 15806.5 (\textit{DQN-PixelCNN}
    \cite{ostrovski2017count}) \\ \hline
    \end{tabular}
\end{center}
\caption{\ours achieves state-of-the-art performance on the hardest games
from the ALE, although the results are not directly comparable, because \ours uses
RAM information not available to \textit{RND} and \textit{DQN-PixelCNN}.
\textit{SOORL} uses even stronger information than \ours.}
\label{tab:perf}
}
\end{table*}

\section{Detailed Setup for Automatically Learning the Abstraction
Function}\label{sec:adm_details}

\begin{table}
\small{
\begin{center}
    \begin{tabular}{ | c | c |}
    \hline
    Envinronment & ADM Accuracy \\ \hline
    \mz & 75\% \\ \hline
    \pitfall & 64\% \\\hline
    \pe & 81\% \\ \hline
    \end{tabular}
\end{center}
\caption{Average ADM accuracy}
\label{tab:ADM}
}
\end{table}

A drawback of \ours is that it uses RAM information to construct the
abstraction function.
Specifically, in all three games, the abstraction function uses the following
information from the RAM:
(i) the agent's location,
(ii) the objects the agent has picked up,
and (iii) the agent's current room number.
The objects that the agent has picked up can be inferred from the cumulative
episodic reward.
The agent's current room number can be inferred by clustering frames by visual
similarity \citep{choi2018contingency} and treating each cluster as a
different room.
Multiple rooms may look the same, and
therefore end up in the same cluster, but they can be disambiguated if the
agent's location can be accurately predicted, by looking at the total distance
traveled by the agent.
When the agent arrives at different rooms, it will have traveled a different
amount to get there.

Thus, the main information our approach requires from the RAM is the agent's
location.
The Attentive Dynamics Model (ADM) from \citet{choi2018contingency}
automatically learns to extract the agent's location via self-supervised
inverse-model loss function.
Using this would enable our approach to operate directly on pixels, without
additional RAM information.
We summarize ADM here, but defer to \citet{choi2018contingency} for the full
details:
ADM applies a 9 x 9 grid over the screen and learns an inverse dynamics model
from each grid cell (predicting the action taken between consecutive frames).
Then, it learns an attention mechanism to place attention on the grid cells
that best predict the agent's action.
These cells are exactly the grid cells containing the agent, because, for
example, predicting that the agent took the left action can only be done from
the cells containing the agent.
We use ADM with the following modifications:

\begin{enumerate}
  \item The attention weights are obtained with softmax instead of sparsemax
    \citep{martins2016softmax} for implementation simplicity.
  \item We found that that the entropy regularization loss $\mathcal{L}_{ent}$
    did not help in learning, so we do not use that term.
  \item We added a temporal consistency loss term $\mathcal{L}_{c}$ to the
    objective with weight 0.1 to encourage the attention to be similar
    across consecutive frames. Specifically, at timestep $t$, $\mathcal{L}_{c}
    = ||\alpha_t - \alpha_{t - 1}||_2$, where $\alpha_t$ is the current
    attention and $\alpha_{t - 1}$ is the attention in the previous timestep.
\end{enumerate}

We evaluate the position extracted by our modified ADM on \mz, \pitfall, and
\pe, and compare it to the actual position of the agent, extracted finding the
cells containing the hard-coded colors matching the agent's sprite.
To integrate ADM with \ours, everytime \ours encountered a new room, we would
collect some frames of training data in the new room via random exploration
for ADM.
Then, we would train ADM on this data (and the data collected from past rooms)
and then freeze ADM on this room, to always produce the same abstract states
in this room.
We simulate this process by training ADM in a progression of rooms obtained
from \ours trained with the RAM abstract state.
On each encounter of a new room, we collect 10000 frames of
data to train ADM on, and a separate 5000 frames to test ADM on by following
the simple exploration policy $\pi^d$ from \refsubsec{discoverer}.
We report the accuracy of ADM averaged over all the rooms in table \ref{tab:ADM},
counting successes as any time ADM predicts a grid cell containing the agent,
or a grid cell one-away from the agent (an amount of error tolerable by
\ours).

As is, ADM makes too many errors to effectively combine it with \ours.
However, this result is a promising proof-of-concept that it's possible to
automatically learn the abstraction function without RAM information.
Future work that improves the accuracy of ADM may make it possible to directly
combine with \ours.

\section{Additional Results}

\subsection{Skill Sharing}\label{sec:skill_sharing_examples}
\begin{figure}\centering
    \subfloat[]{{\includegraphics[width=0.32\linewidth]{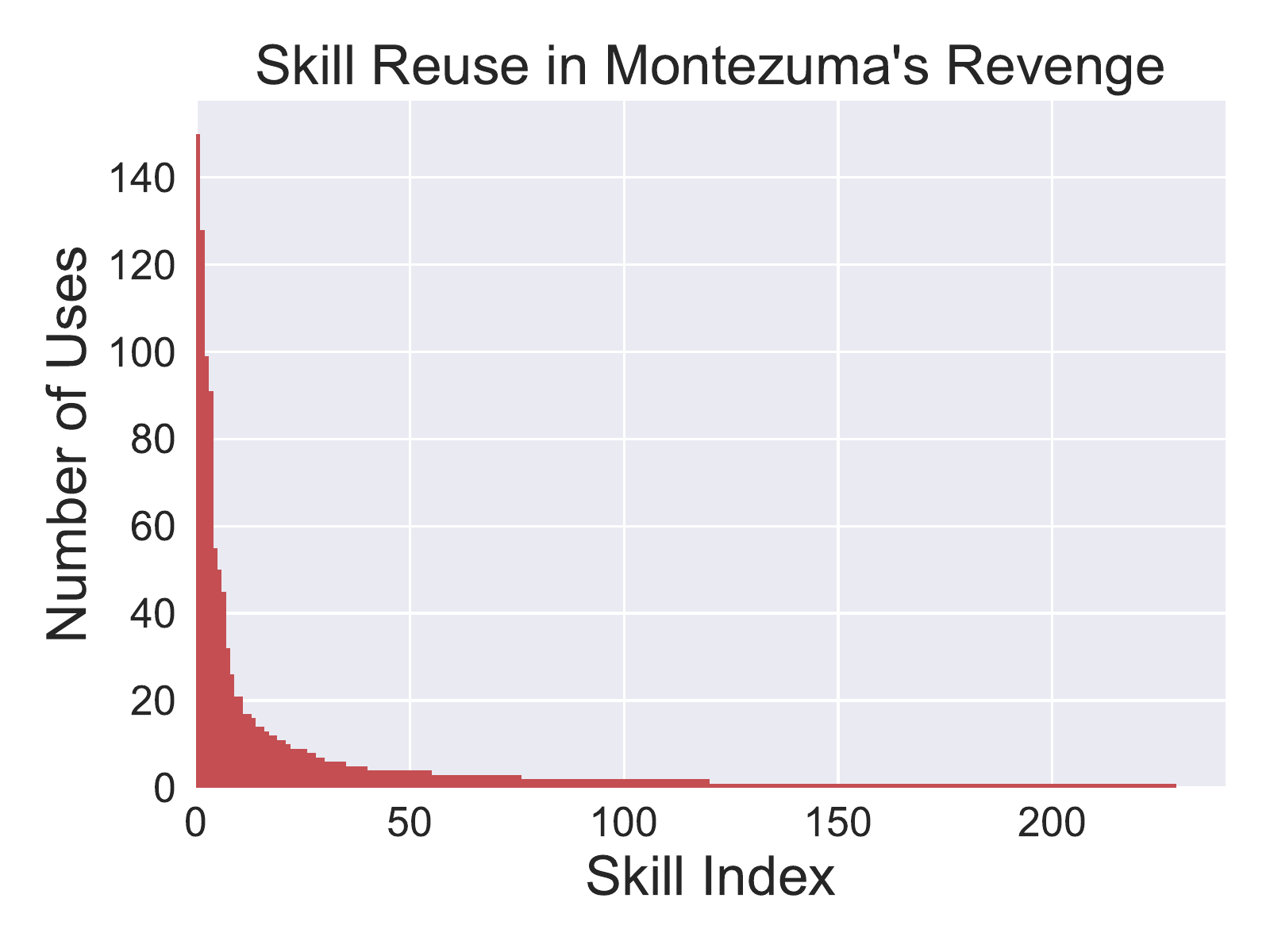} }}\subfloat[]{{\includegraphics[width=0.32\linewidth]{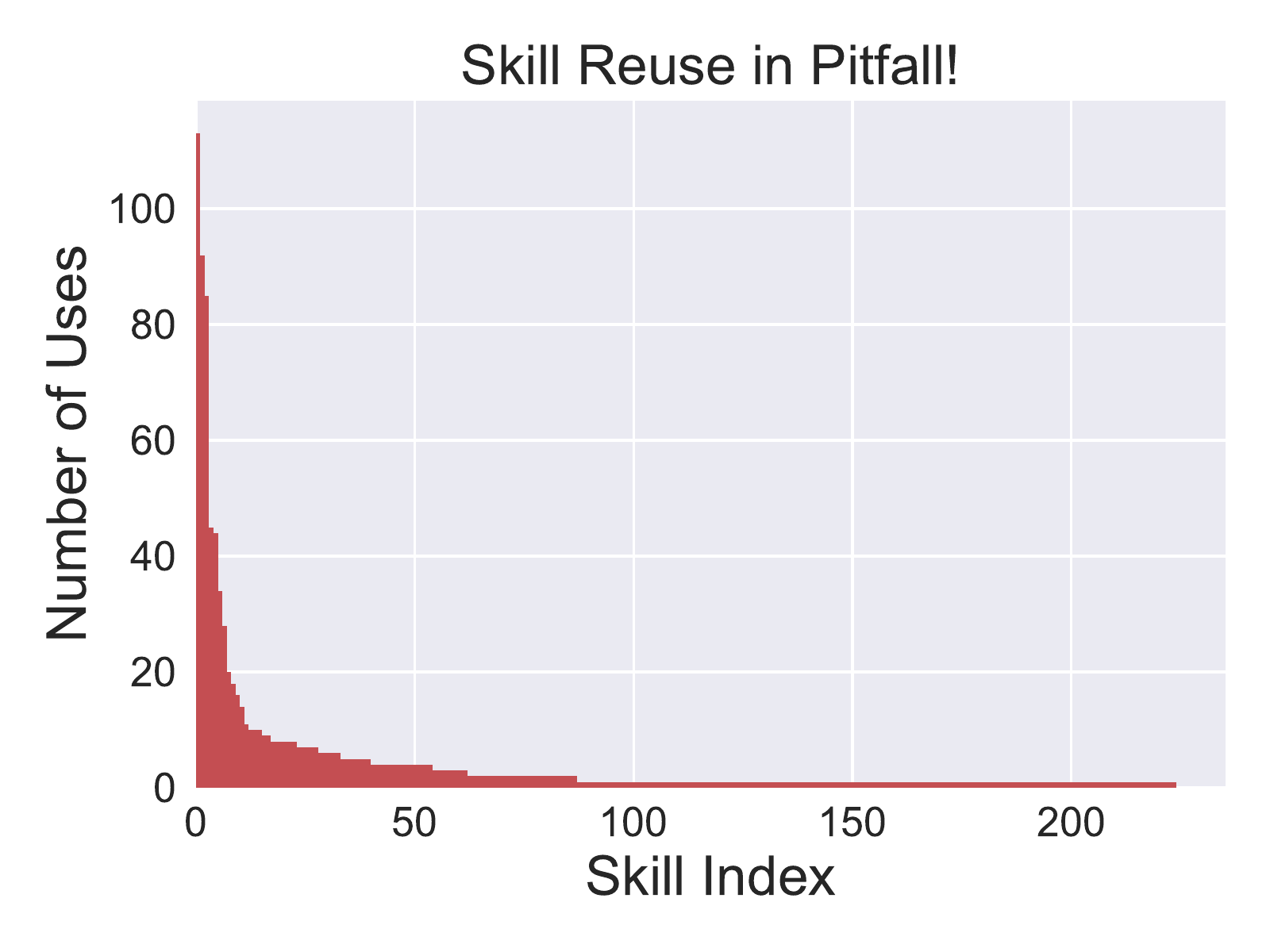} }}\subfloat[]{{\includegraphics[width=0.32\linewidth]{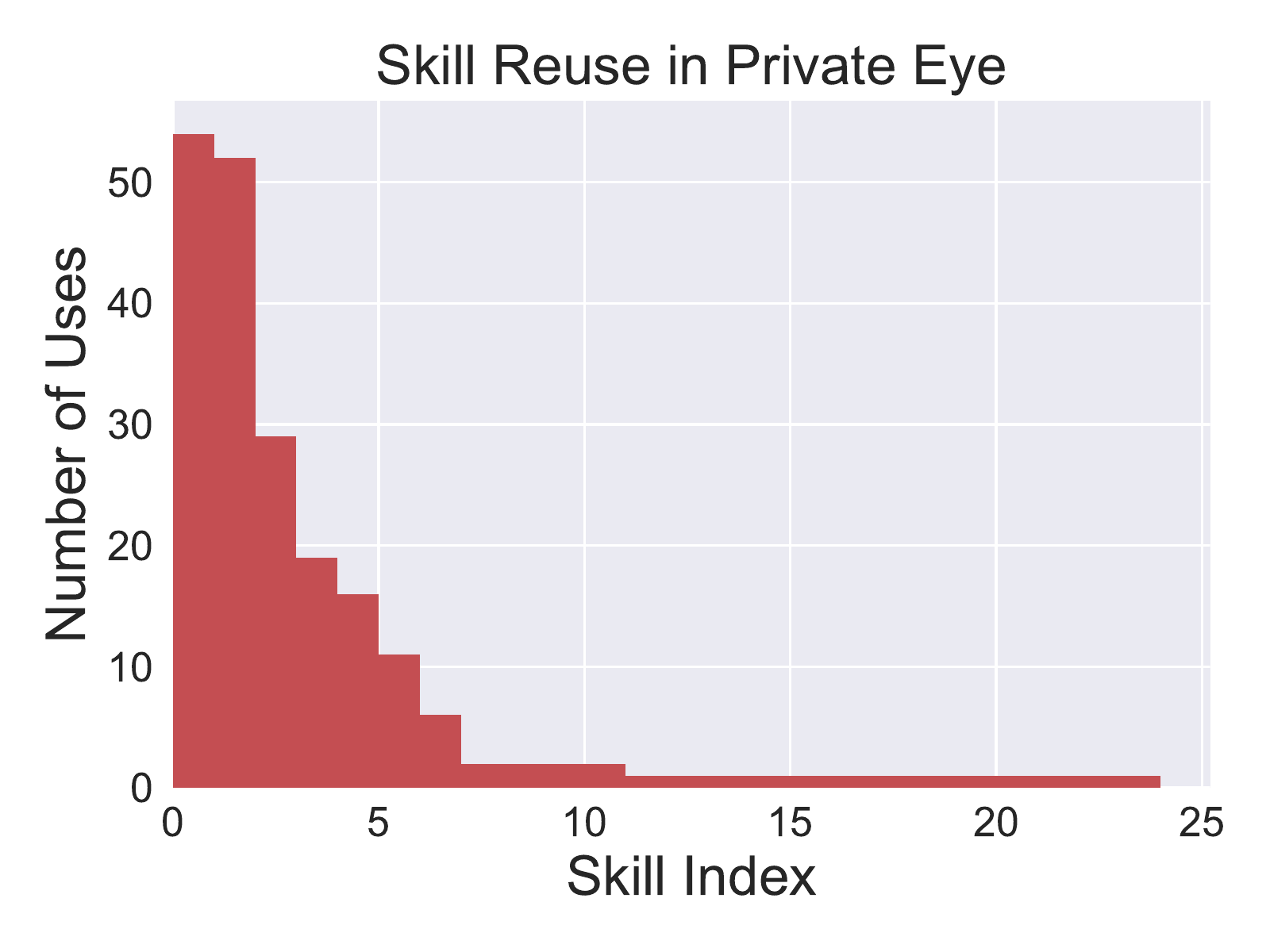} }}\caption{
      The number of different transitions each skill can traverse.
      Skills are sorted by decreasing usage.
    }\label{fig:skill-histogram}\end{figure}

The worker learns skills that successfully apply in to many similar
transitions.
\reffig{skill-histogram} depicts the number of different transitions each
skill is used on in \mz, \pitfall, and \pe.
The simplest skills (e.g. move to the left) enjoy the highest number of reuses,
while more esoteric skills (e.g. jump over a particular monster)
are only useful in few scenarios.

\begin{figure}\centering
    \subfloat[]{
        \includegraphics[width=0.32\linewidth]{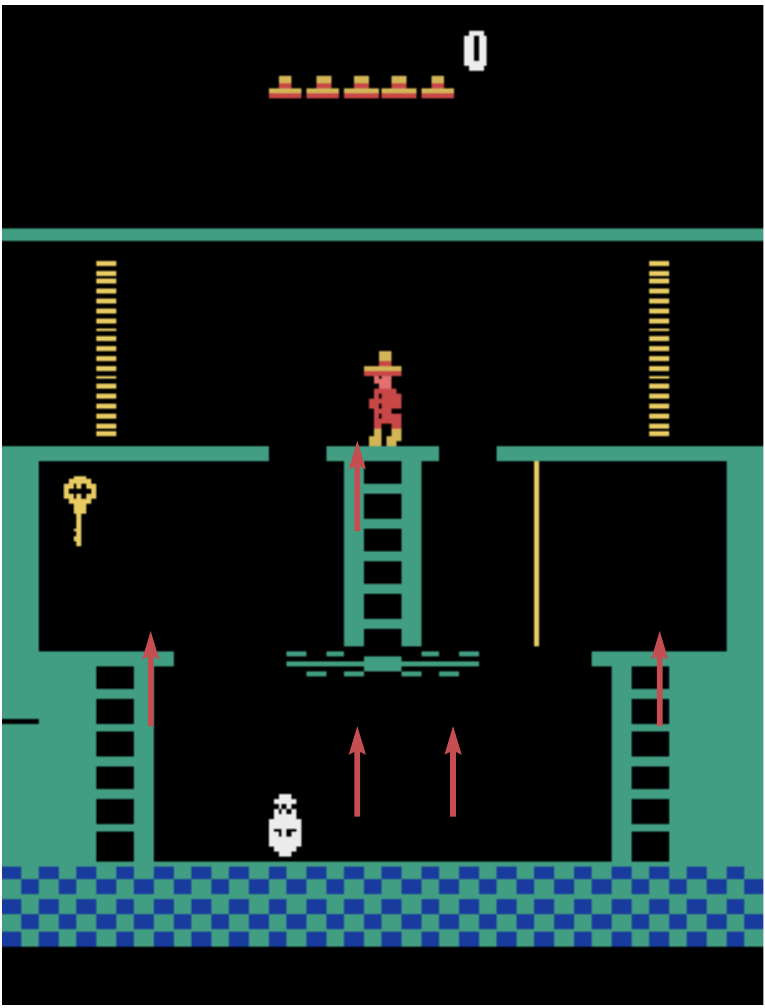}\label{ladder_monster}}
    \subfloat[]{
        \includegraphics[width=0.32\linewidth]{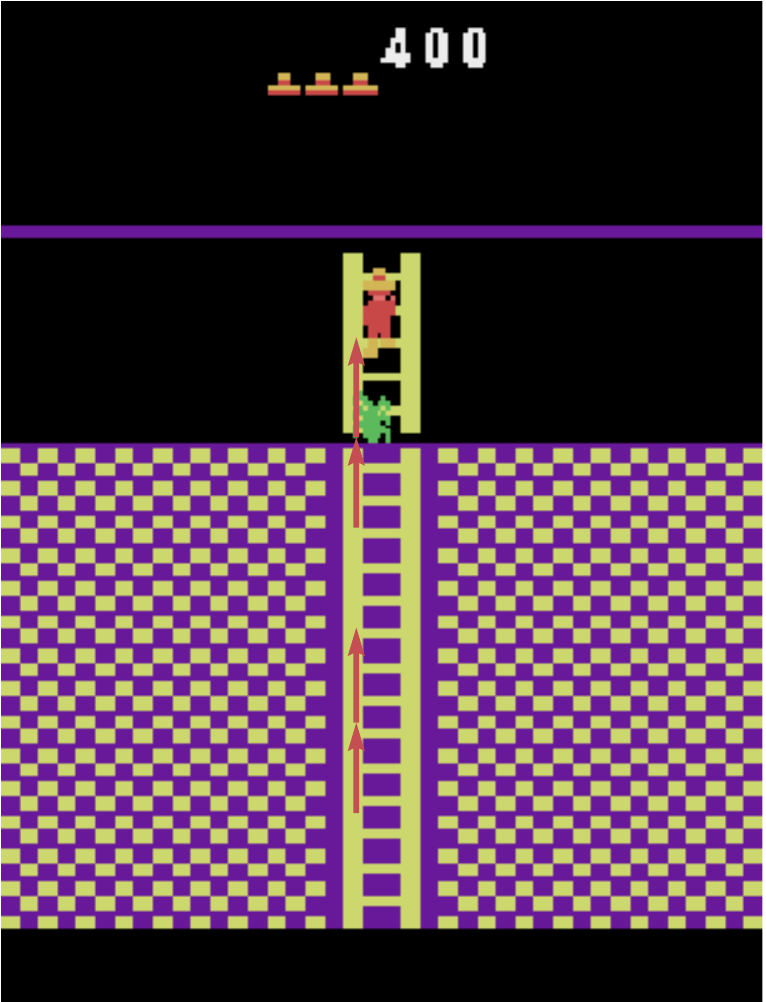}\label{ladder}}
    \caption{
        Skill reuse in \mz.
        The same skill is useful in multiple rooms and can both climb up
        ladders and jump over a monster.
    }\label{fig:skill-share-mz}\end{figure}

\reffig{skill-share-mz}
provides an example of a skill in \mz with relatively high reuse.
The arrows denote the movement of the agent when it executes the skill.
The same skill that jumps over a monster in the first room
(\reffig{skill-share-mz}\subref{ladder_monster}) can also climb up
ladders.
In \reffig{skill-share-mz}\subref{ladder},
the skill appears to know how to climb up all parts of the ladder except for
this middle.
This occurs because the spider occasionally blocks the middle of the ladder,
and a different special skill must be used to avoid the spider.
However,
the skill reuse is not perfect.
For example,
in \reffig{skill-share-mz}\subref{ladder_monster},
the skill can climb up the top half of ladders,
but a separate skill climbs the bottom half of the ladders.

\subsection{Generalization to New Tasks}\label{sec:full_new_rewards}
\begin{figure*}[t]
    \centering
    \includegraphics[width=\linewidth]{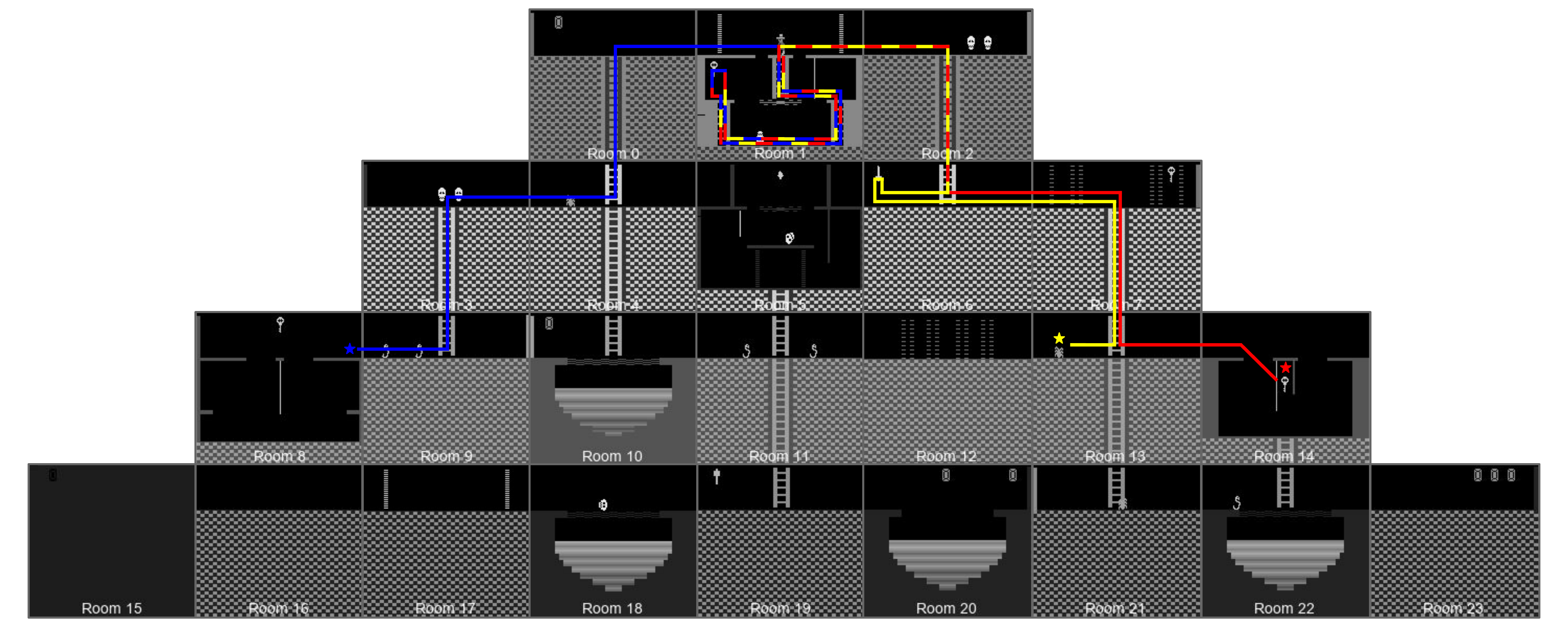}
    \caption{
      Display of all the rooms of the \mz pyramid.
      The agent starts in room 1 and must navigate through the pyramid,
      picking up objects and dodging monsters to complete the new tasks.
      The end of each task is marked with a star.
      Example paths for each task are marked with different colors.
      Multiple colors indicate sections of the paths that are shared across
      multiple tasks.
      (Red: \textit{Get key}, Yellow: \textit{Kill spider}, Blue:
      \textit{Enter room 8}).
    }
    \label{fig:pyramid}
\end{figure*}

\begin{figure}\centering
    \subfloat[]{{\includegraphics[width=0.32\linewidth]{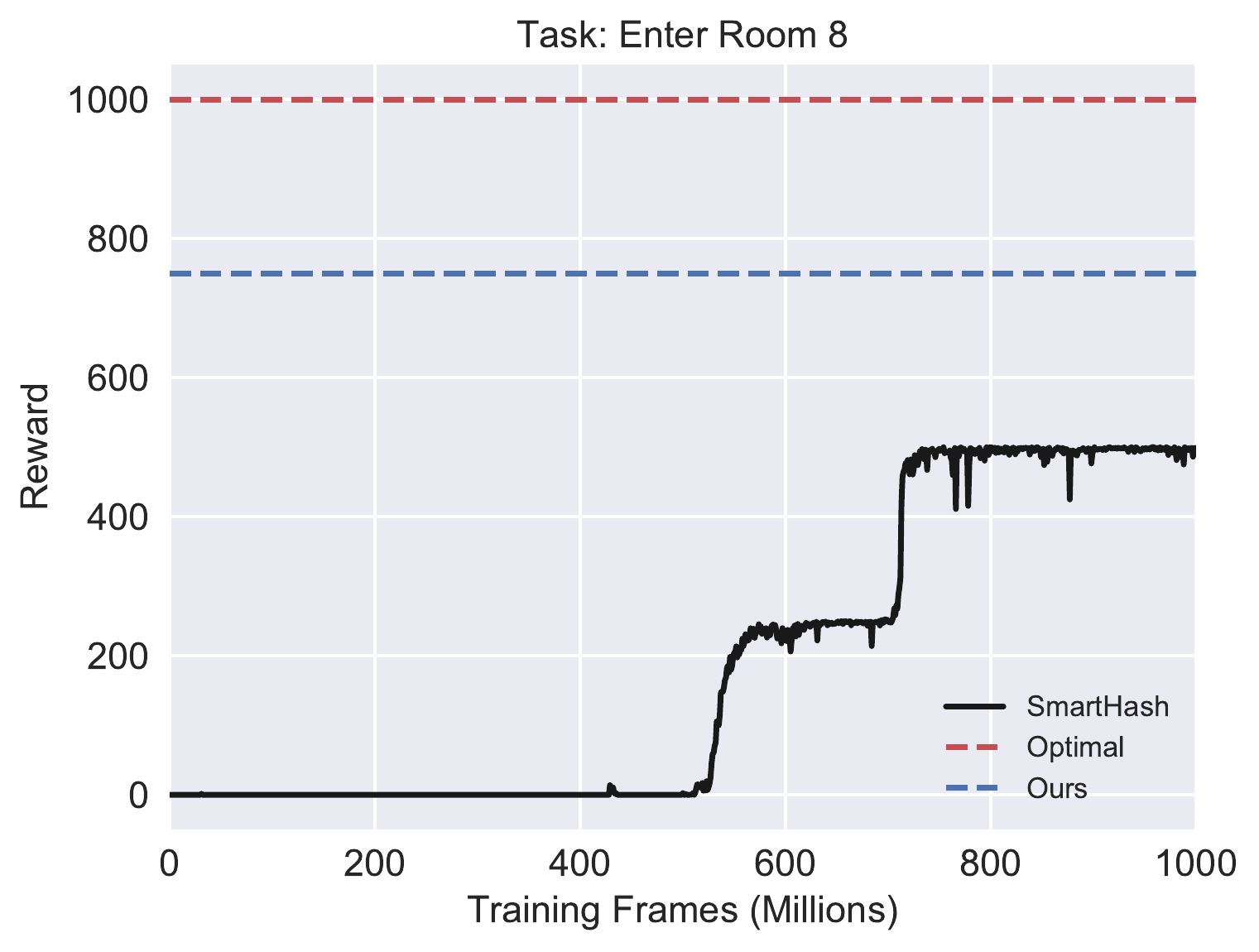} }}\subfloat[]{{\includegraphics[width=0.32\linewidth]{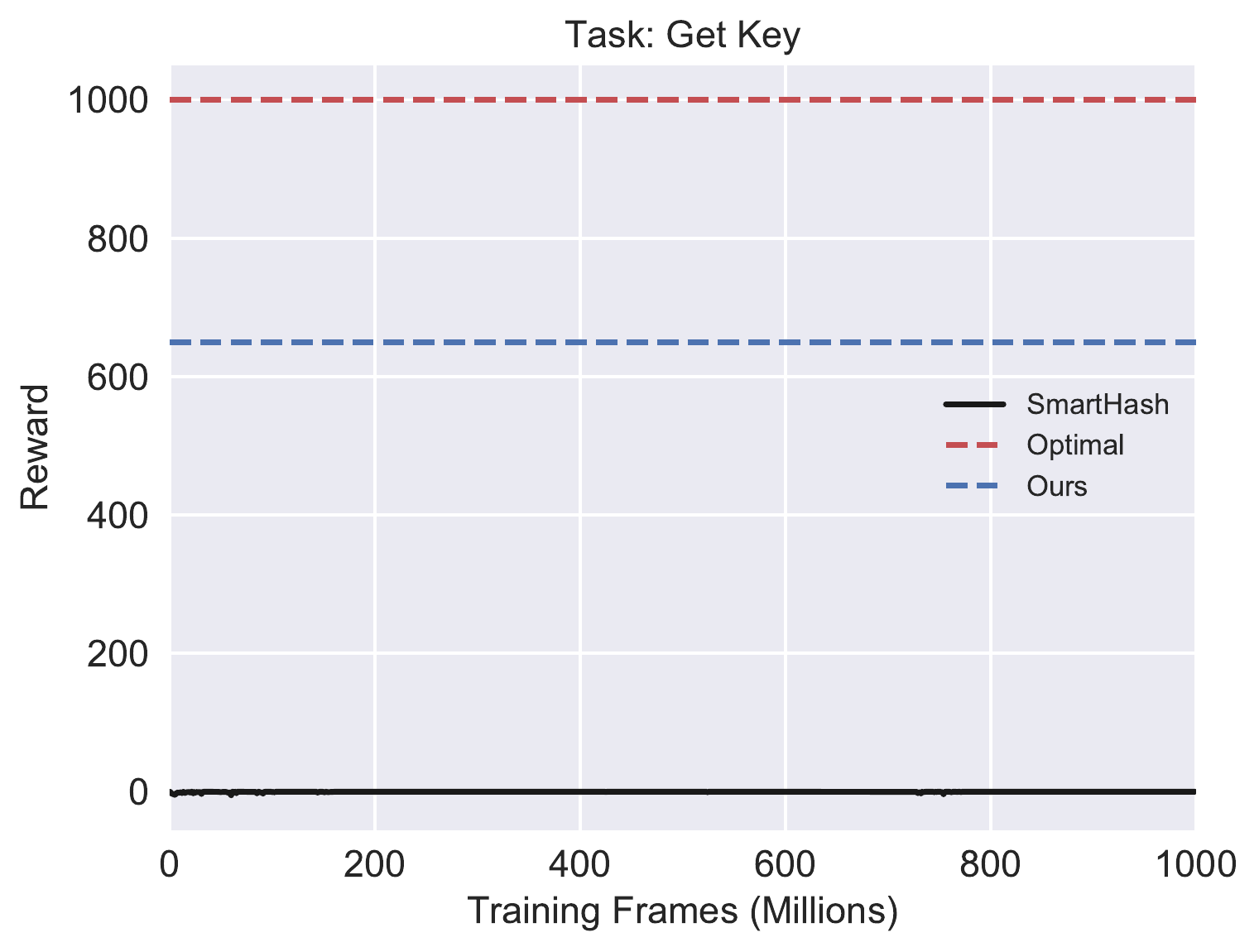} }}\subfloat[]{{\includegraphics[width=0.32\linewidth]{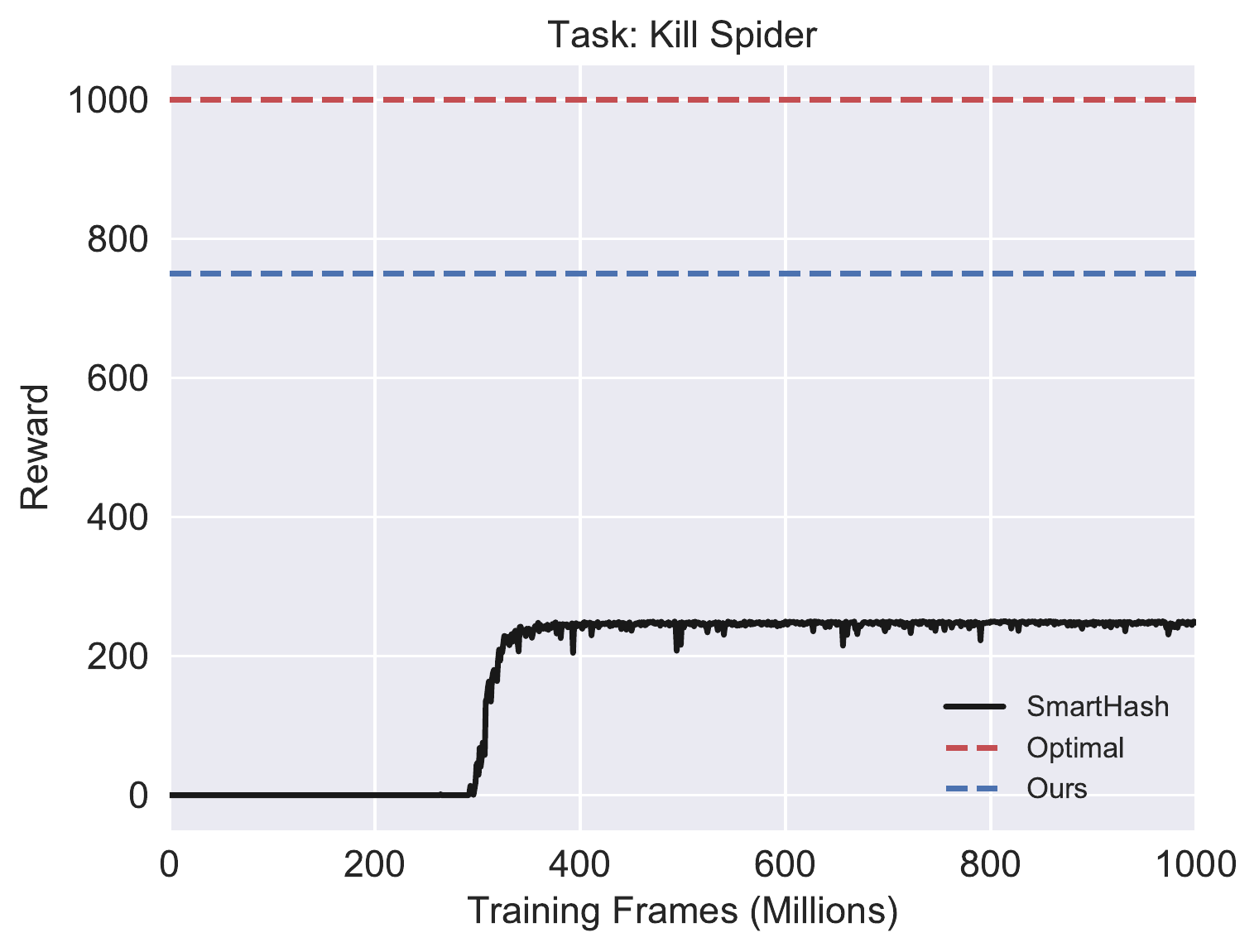} }}\caption{
    Training curves of \textit{SmartHash} on other reward functions in \mz,
    compared with the performance of our approach generalizing to the new task.
    Our approach quickly transfer with 1M frames, whereas \textit{SmartHash}
    trains from scratch for 1000x more frames, and still achieves lower
    reward.
    }\label{fig:new_rewards}\end{figure}

To evaluate the ability of our approach to generalize to new reward functions,
we train our approach on the basic \mz reward function and then test it on
three challenging new reward functions (illustrated in \reffig{pyramid}),
not seen during training:

\begin{itemize}
  \item \emph{Get key:} the agent receives 1000 reward for picking up the key
    in room 14 (6 rooms away from the start).
    In addition,
    the agent receives -100 reward for picking up any other objects or opening
    any other doors.
  \item \emph{Kill spider:} the agent receives 1000 reward for killing the
    spider in room 13 (5 rooms away from the start).
    To kill the spider,
    the agent must first pick up the sword in room 6
    (3 rooms away from the start)
    and save the sword for the spider.
    The agent receives no other reward.
  \item \emph{Enter room 8:} the agent receives 1000 reward for entering room
    8 (6 rooms away from the start).
    The agent receives no other reward.
\end{itemize}

In all three tasks,
the episode ends when the agent completes its goal and receives positive
reward.

Our approach trains on the basic \mz reward function for 2B frames,
and then is allowed to observe the new reward functions for only 1M frames.
We compare with \textit{SmartHash},
which trains directly on the new reward functions for 1B frames.
The results are summarized in \reffig{new_rewards}.
Even when evaluated on a reward function different from the reward function it
was trained on,
our approach achieves about 3x as much reward as \textit{SmartHash},
which is trained directly on the new reward function.
Averaged over all 3 tasks,
our approach achieves an average reward of $716.7$ out of an optimal reward of
$1000$,
whereas \textit{SmartHash} only achieves an average reward of $220$,
even when trained directly on the new reward function.
These experiments suggest that after our approach is trained in an environment
on one task,
it can quickly and successfully adapt to new tasks with little additional
training.

\subsection{Near-Linear Training}\label{sec:linear}
\begin{figure}\centering
    \subfloat[]{{\includegraphics[width=0.32\linewidth]{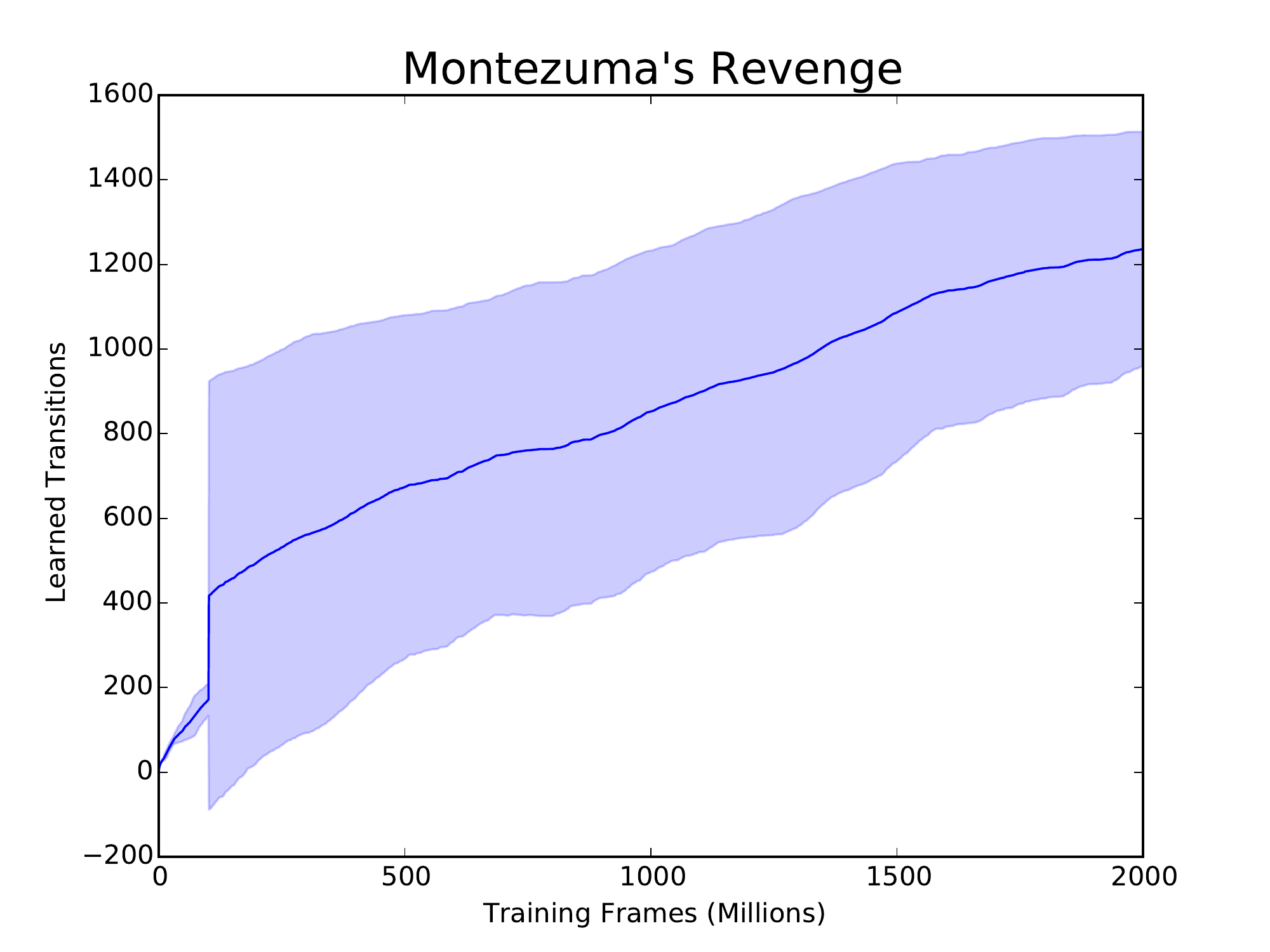} }}\subfloat[]{{\includegraphics[width=0.32\linewidth]{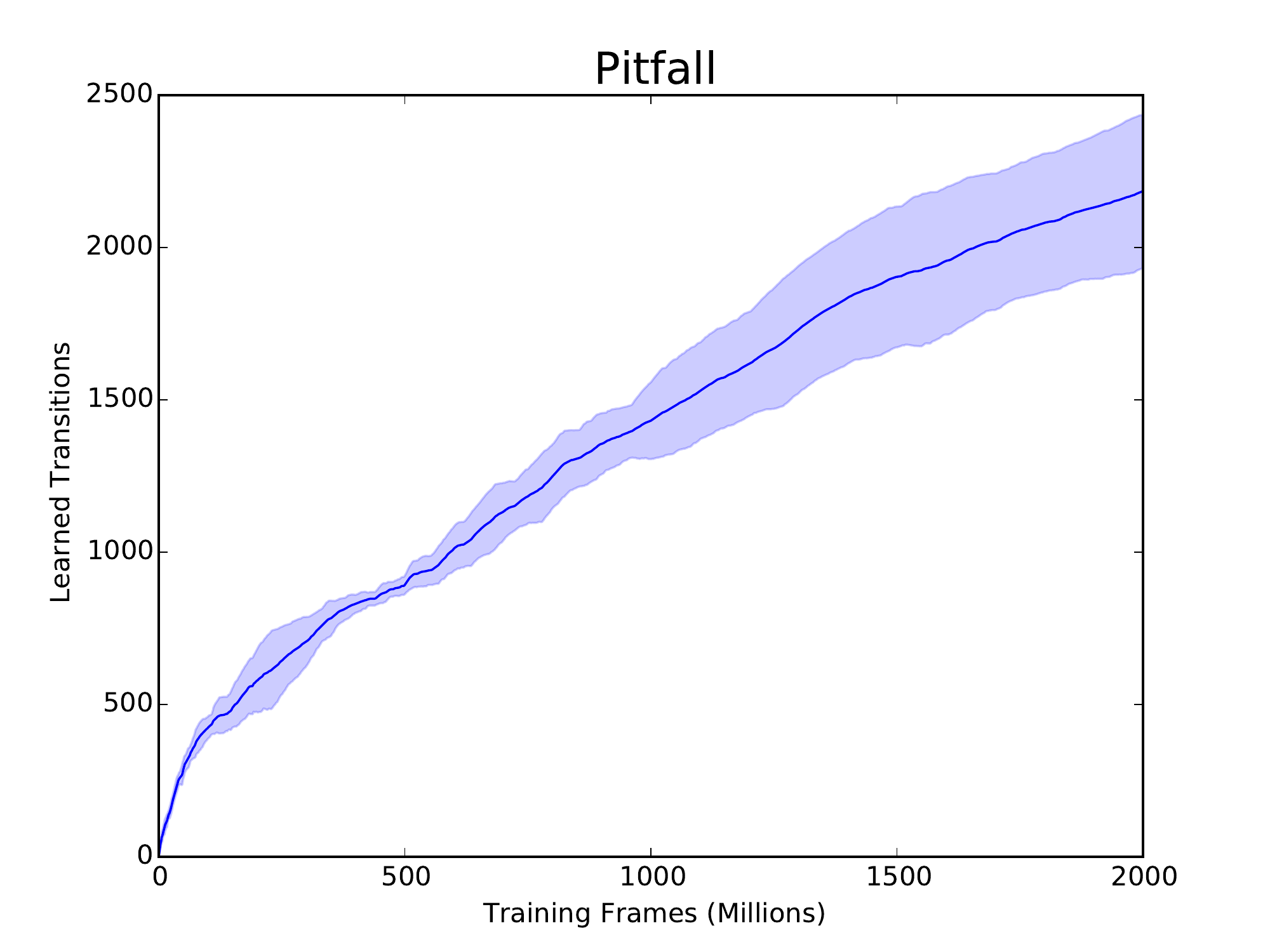} }}\subfloat[]{{\includegraphics[width=0.32\linewidth]{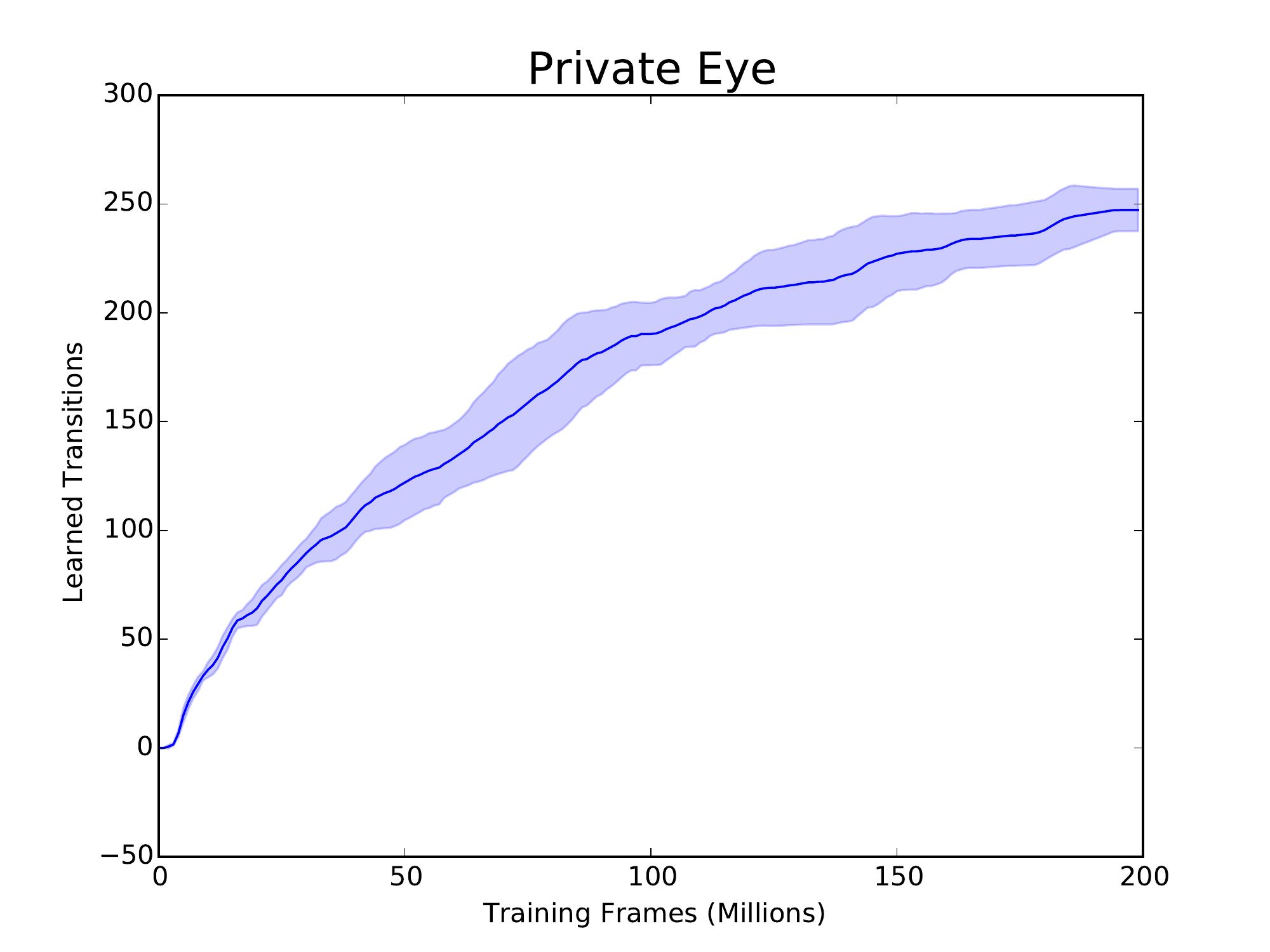} }}\caption{
      The number of transitions learned by the worker vs. number of training
      frames.
      The worker continues to learn new transitions even late into training,
      showing almost no signs of slowing down in \mz and \pitfall.
    }\label{fig:transitions}\end{figure}

Whereas many prior state-of-the-art approaches tend to plateau toward the end
of training,
our approach continues to make near-linear progress.
\reffig{transitions} graphs the number of transitions learned by the worker
against the number of frames in training.
In \mz and \pitfall particularly,
the rate the worker learns new transitions is nearly constant throughout
training.
Because of this, when we continued to train a single seed on \pitfall, by 20B
frames, it achieved a reward of 26000, and by 80B frames, it achieved a reward
of 35000.

\subsection{Additional Results on \pe}\label{sec:private_eye}

\begin{figure}\centering
    \subfloat[]{{\includegraphics[width=0.45\linewidth]{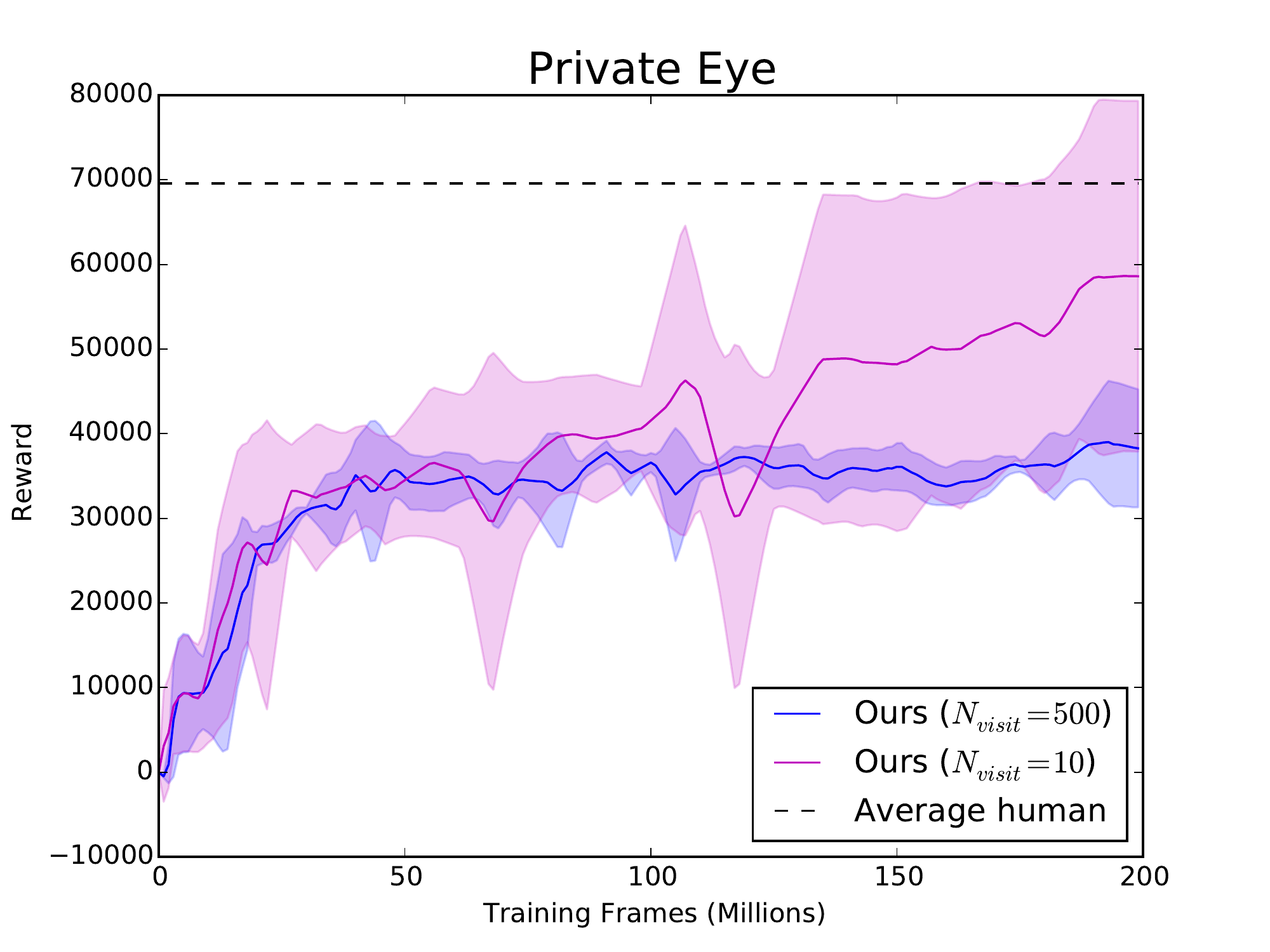}}}\subfloat[]{{\includegraphics[width=0.45\linewidth]{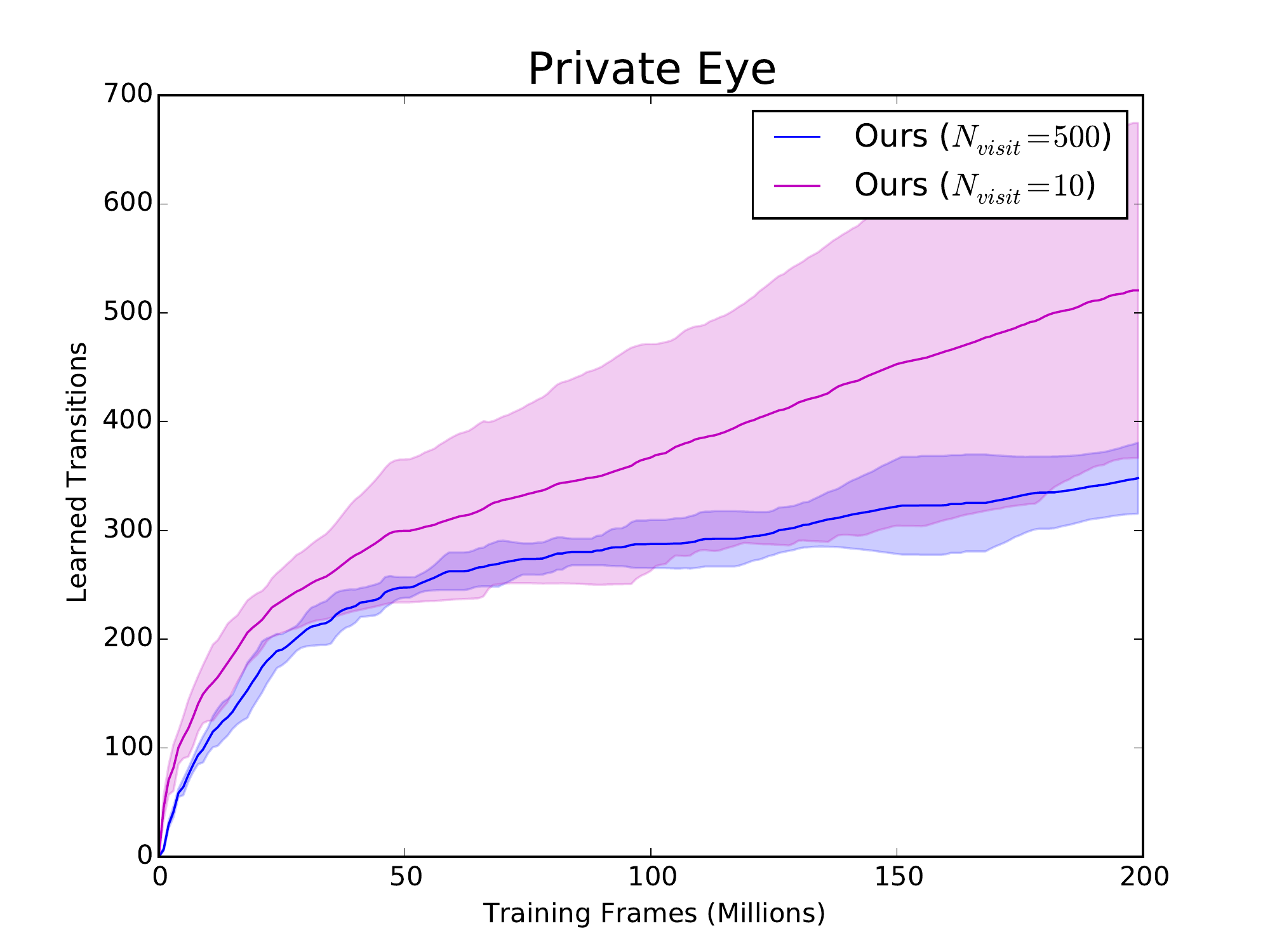} }}\caption{
      Performance of our approach on \pe,
      when decreasing the number of exploration episodes ($N_{visit}$).
      (a)
      When $N_{visit}$ is set to 10,
      our approach performs even better,
      achieving near human-level rewards.
(b)
      Decreasing $N_{visit}$ enables the worker to learn more transitions
      in fewer frames.
    }\label{fig:private_eye_tuned}\end{figure}

By changing a single hyperparameter,
our approach can perform even better on \pe,
exceeding human performance on 2 of 4 seeds.
Since nearby abstract states are particularly easy to discover in \pe,
the manager needs not explore for new transitions as many times.
Consequently,
if we decrease the number of times the manager explores around each
abstract state $N_{visit}$ from the 500
(used in the main experiments for all three games)
to 10,
performance improves.
\reffig{private_eye_tuned} compares the performance with the decreased value of $N_{visit}$
with the original value of $N_{visit}$ reported in the main experiments.
Decreasing $N_{visit}$ prevents the manager from wasting frames with
unnecessary exploration and consequently enables the worker to learn more
transitions in fewer total frames.
With $N_{visit}$ set to 10,
our approach achieves a final average performance of 60247 after 200M frames
of training.
Additionally,
the top 2 of our 4 seeds achieve rewards of 75600 and 75400,
exceeding average human performance: 69571 \citep{pohlen2018observe}.

\section{Additional Seeds}
\begin{figure*}\centering
    \subfloat[]{{\includegraphics[width=0.30\linewidth]{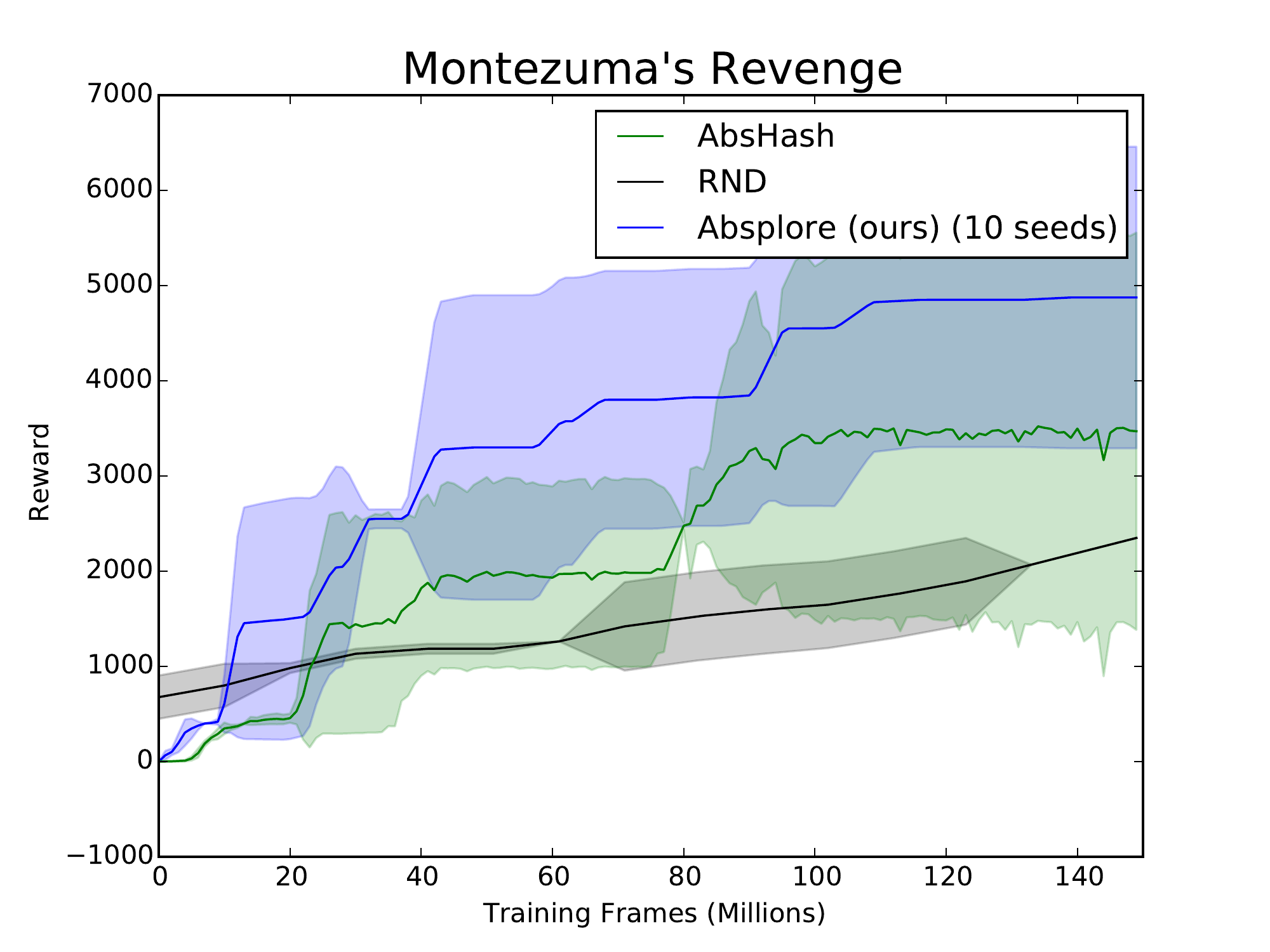} }}\subfloat[]{{\includegraphics[width=0.30\linewidth]{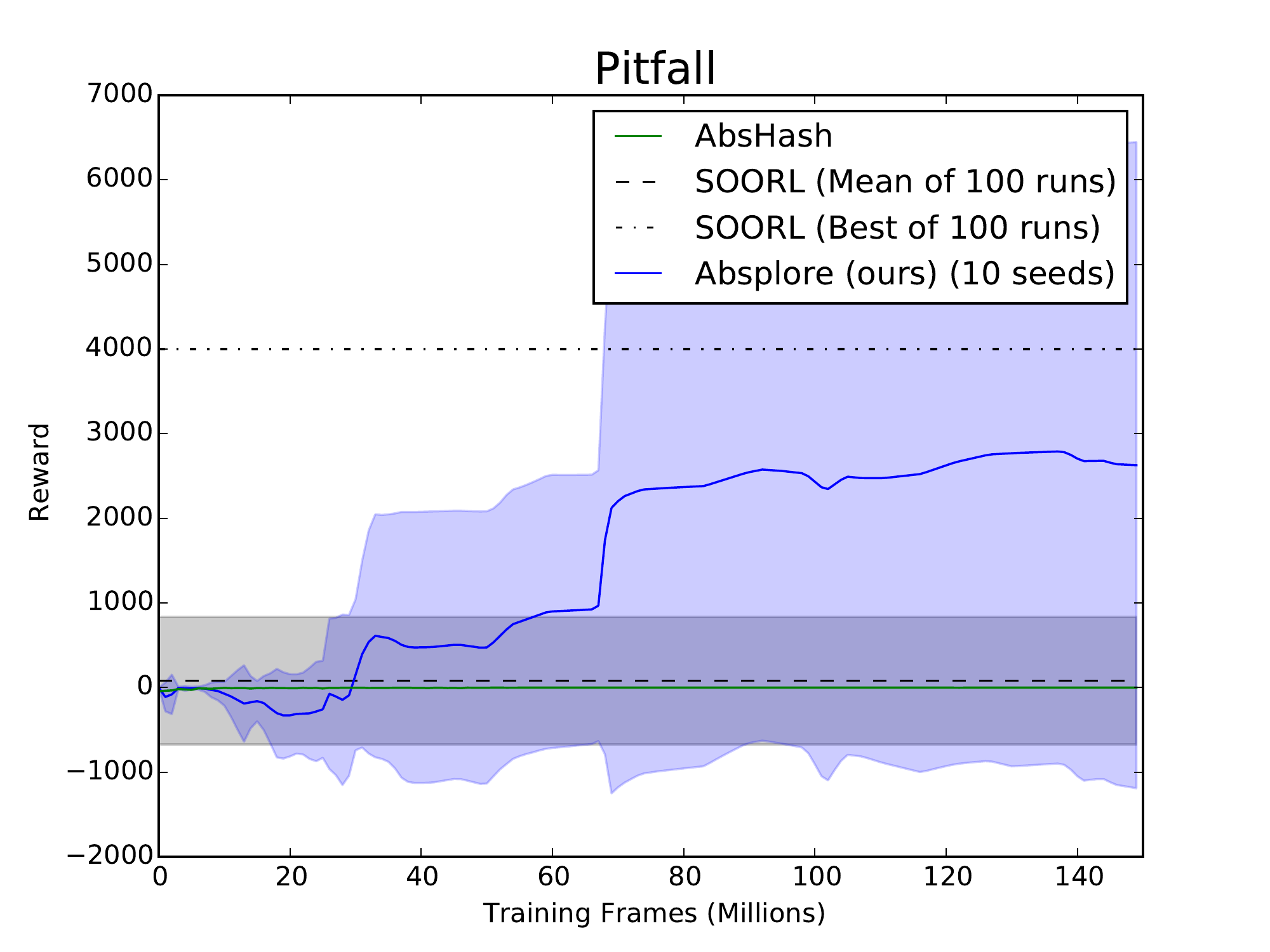} }}\subfloat[]{{\includegraphics[width=0.30\linewidth]{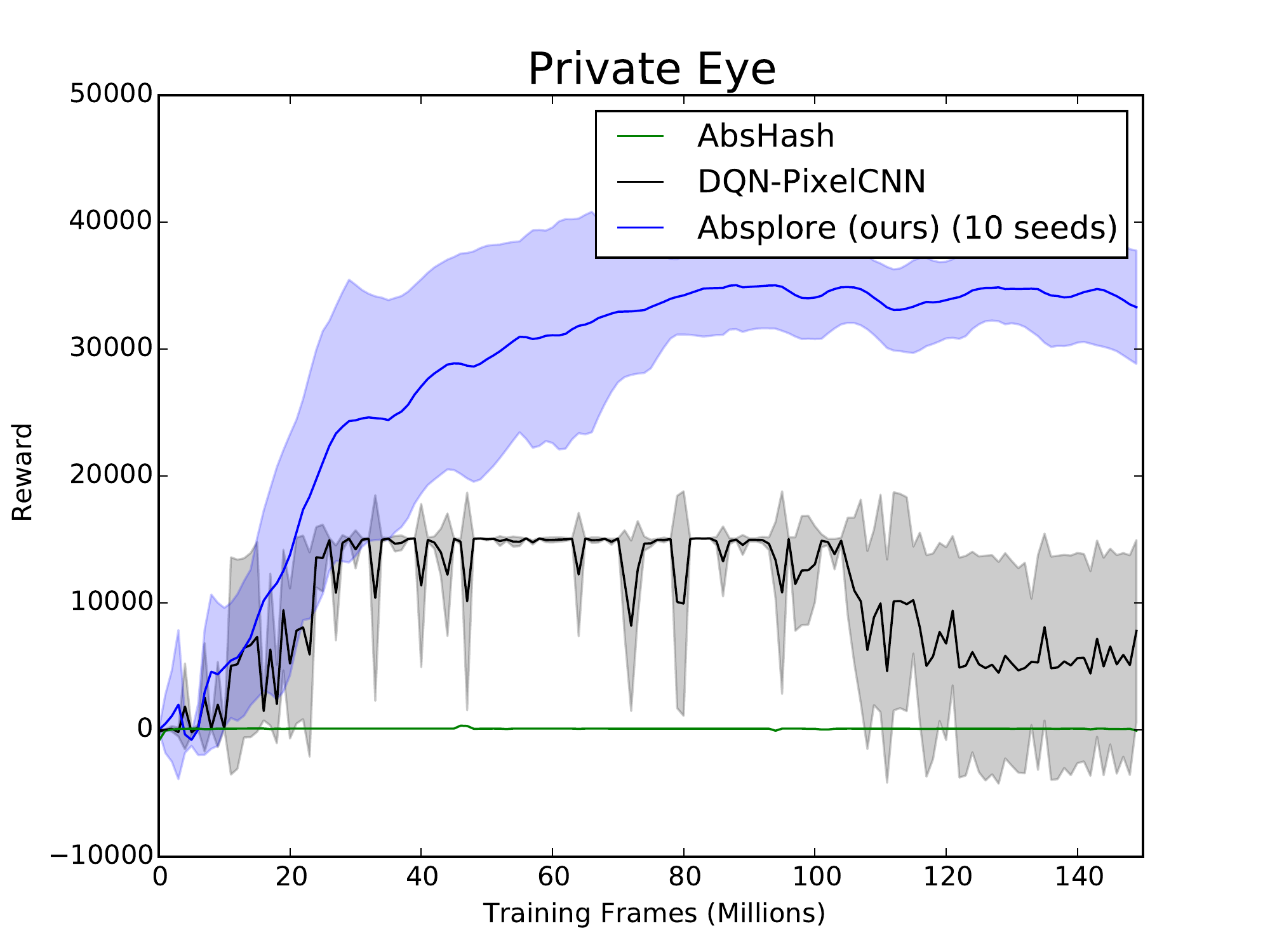} }}\caption{
      Stability:
      Comparison of \ours with the prior non-demonstration state-of-the-art
      approaches in \mz, \pitfall, and \pe.
      \ours is averaged over 10 different random seeds.
    }\label{fig:stable}\end{figure*}

\cite{henderson2017deep} emphasizes the importance of running many seeds to
combat the high variance of deep RL results.
We report additional results with 10 different random seeds of \ours on \mz,
\pitfall, and \pe in \reffig{stable}.
Due to computational constraints, we only run for 150M frames, but even with
more seeds, our approach still compares favorably to the prior
state-of-the-art.

\section{Guarantees for Near Optimality}\label{sec:proofs}

In general,
a hierarchical policy over skills is not guaranteed to be near-optimal,
because certain optimal trajectories may be impossible to follow using the
skills.
Because of this,
hierarchical reinforcement learning literature typically focuses on
hierarchical optimality \citep{dietterich2000hierarchical}
optimality given the abstractions.
However, under the following assumptions, our approach provably achieves a
near-optimal policy on the original MDP in time polynomial in the size of the
abstract MDP, with high probability.

\paragraph{Notation.}
Recall that we are interested in finding a near-optimal policy on the concrete
MDP, with concrete states $x \in \mathcal{X}$ and concrete actions $a \in
\mathcal{A}$.
We refer to the value function of the optimal policy in the concrete MDP as
$V^*(x)$.

From the concrete MDP approach constructs the abstract MDP, consisting of
abstract states $s$ in the known set $\mathcal{S}$, learned abstract
actions $o$ (e.g., $\text{go}(s, s')$), transition dynamics $P(s' | o, s)$,
and rewards $R(s, s')$.
The abstract MDP changes over time.
We refer to the known set at timestep $t$ as $\mathcal{S}_t$ and to the set of
all abstract states as $\Phi = \{\abstraction(x) : x \in \mathcal{X}\}$.
We refer to the optimal value function on the abstract MDP at
timestep $t$ as $V^*_t(s)$.

Our approach maintains estimates of the rewards and transition
dynamics of the abstract MDP.
With these estimates, at each timestep $t$, our approach computes $\pi_t$, the
policy that is optimal with respect to these models (e.g., via value
iteration).
To simplify notation, we refer to the expected reward achieved by $\pi_t$ on
the abstract MDP at timestep $t$ starting at abstract state $s$ as
$V^{\pi_t}_t(s)$.
The policy computed by our approach, $\pi_t$ also applies on the concrete MDP,
because actions on the abstract MDP are implemented as subpolicies on the
concrete MDP.
Consequently, we refer to the expected reward achieved by $\pi_t$ on the
concrete MDP starting at concrete state $x$ as $V^{\pi_t}(x)$.

\paragraph{Assumptions.}
Formally, we require the following assumptions:

\begin{enumerate}
  \item
    The learned abstract MDP is deterministic.

  \item
    The learned abstract MDP has rewards that are \emph{path independent}:
    i.e., all trajectories to an abstract state $s$ achieve the same
    reward.
  \item
    The diameter of each abstract state is at most $H_{worker}$,
    where we define the diameter of an abstract state $s$ to be the maximum
    number of steps required to navigate to an immediate neighbor $s'$.
\end{enumerate}

Assumption 1 intuitively says that the worker can successfully abstract away
stochasticity from the manager, which our experiments in
\refsec{stochasticity} suggests is possible.
Humans typically also make this assumption when they plan.
For example, when humans plan (e.g., to get to Paris), they expect to
deterministically hit subgoals (e.g., get to the airport, get on the plane,
get to the hotel) even though the world is actually non-deterministic (e.g.,
the taxi may be late.)

Assumption 2 tends to hold under many natural abstraction functions.
For example, in the ALE games we evaluate on, the state abstraction function
captures the agent's inventory and a history of the agent's inventory.
Since all reward in these games is given when the agent picks up new items, or
uses an item in its inventory, the agent's inventory and history encodes path
independent reward.
This also holds for many robotic arm manipulation tasks.
For example, in a block stacking task with sparse rewards, a natural state
abstraction might be the location of all the blocks.
Then, the reward of a trajectory is encoded by the last abstract state of the
trajectory: all trajectories that lead to a stacked configuration of blocks
achieve the same reward for a success, while all non-stacking trajectories
achieve the same failure reward.

Assumption 3 ensures that the worker has enough timesteps to navigate to
any immediate neighbors.
This is easily satisfied by setting $H_{worker}$ conservatively.

\paragraph{Main results.}
While the above assumptions enable us to prove near-optimality,
our method performs well empirically even when these assumptions are violated.
Given the above assumptions,
our main theoretical result holds:

\begin{proposition}\label{prop:main}
  Under the assumptions, for a given input $\eta$ and $\epsilon$,
  $\pi_t$ is at most $\epsilon$ suboptimal,
  $V^{\pi_t}(x_0) \geq V^*(x_0) - \epsilon$, on all
  but the first $O\big(|\Phi|^3 (|\mathcal{A}| +
  \frac{\log{K|\Phi|} +
  \log{\frac{1}{\eta}}}{\log{\frac{1}{p}}}) + d_{max} \times H_{worker}\big)$ timesteps,
  where $x_0$ is the starting concrete state
  and $p$ and $K$ are polynomial in $|\Phi|$ and $|\mathcal{A}|$.
\end{proposition}

To prove \refprop{main},
we require the following three lemmas:

\begin{lemma}
  \label{lem:amdp_optimality}
  By setting $N_{transition}$ to be
  $O(\frac{\log{1 - (1 - \eta')^{\frac{1}{|\Phi|^2}} / 2}}{2(\epsilon / |\Phi|HV^*(x_0))^2})$,
  with probability $1 - \eta'$,
  at each timestep $t$,
  $\pi_t$ is near-optimal on the current abstract MDP:
  i.e., $V^{\pi_t}_t(s) \geq V^*_t(s) - \epsilon$ for all
  abstract states $s \in \Phi$.
\end{lemma}

\begin{lemma}
  \label{lem:optimality_transfer}
  If the known set is equal to the set of all
  abstract states ($\mathcal{S} = \Phi$) at timestep $T$,
  then for any policy $\pi$ on the abstract MDP, $\pi$ achieves the same
  expected reward on the abstract MDP as on the concrete MDP:
  i.e., $V^\pi_T(\abstraction(x_0)) = V^\pi(x_0)$, where $x_0$ is the initial
  concrete state.

  In addition, the expected return of the optimal policy on the abstract MDP
  is equal to the expected return of the optimal policy on the concrete MDP:
  i.e., $V^*_T(\abstraction(x_0)) = V^*(x_0)$ where $x_0$ is the initial
  state.
\end{lemma}

\begin{lemma}
  \label{lem:number_samples}
  With probability $1 - \eta$,
  the known set grows to cover all abstract states in
  $O\big(|\Phi|^3 (|\mathcal{A}| + \frac{\log{K|\Phi|} +
  \log{\frac{1}{\eta}}}{\log{\frac{1}{p}}}) +
  d_{max} \times H_{worker}\big)$ time.
\end{lemma}

Given these lemmas,
we are ready to prove \refprop{main}:

\begin{proof}[Proof of \refprop{main}]
  For simplicity,
  we ignore terms due to appropriately setting $N_{transition}$ and
  $1 - \eta'$ from \reflem{amdp_optimality},
  but these terms are all polynomial in the size of the abstract MDP.

  By \reflem{number_samples}
  the known set grows to cover all abstract states in
  $T = O\big(|\Phi|^3 (|\mathcal{A}| + \frac{\log{K|\Phi|} +
  \log{\frac{1}{\eta}}}{\log{\frac{1}{p}}}) + d_{max} \times H_{worker}\big)$
  timesteps.
  For all timesteps $t \geq T$, by \reflem{amdp_optimality}, $\pi_t$ is at
  most $\epsilon$ suboptimal on the abstract MDP.
  On all those timesteps, the known set is equal to all abstract states,
  so by \reflem{optimality_transfer}, $\pi_t$ is at most $\epsilon$ suboptimal
  on the concrete MDP.
\end{proof}

\paragraph{Proofs.}
Now,
we prove \reflem{amdp_optimality}, \reflem{optimality_transfer}, and
\reflem{number_samples}.

\begin{proof}[Proof of \reflem{amdp_optimality}]
  Let $\hat{P}(s' | o, s)$ denote the estimated transition dynamics\footnote{In \refsec{growing}, we simplify notation to estimate the
    success rate instead of the full dynamics, but the manager could have
    estimated the full dynamics as required here.}
  and
  $\hat{R}(s, s')$ denote the estimated reward model in the abstract MDP.

  For each reliable transition $(s, s')$ (action in the abstract MDP), the
  manager estimates $\hat{P}(s' | o, s)$ from $N_{transition}$ samples of the
  worker.
  We bound the error in the model $|\hat{P}(s' | o, s) - P(s' | o, s)|$ with high
  probability by Hoeffding's inequality:

  \begin{equation}
    P(|\hat{P}(s' | o, s) - P(s' | o, s)| \geq \alpha) \leq 2e^{-2N_{transition}\alpha^2}
  \end{equation}

  By the Assumption 2, $\hat{R}(s, s') = R(s, s')$ because all trajectories
  leading to $s$ achieve some reward $r$ and all trajectories leading to $s'$
  achieve some reward $r'$, so a single estimate $R(s, s') = r' - r$ is
  sufficient to accurately determine $\hat{R}$.

  Because the model errors are bounded, and because the abstract MDP is
  Markov, we can apply the simulation lemma \citep{kearns2002near},
  which states that if $|\hat{P}(s' | o, s) - P(s' | o, s)| \leq \alpha$ and
  $|\hat{R}(s, s') - R(s, s')| \leq \alpha$,
  then the policy $\pi$ optimizing the MDP formed by $\hat{P}$ and $\hat{R}$
  is at most $\epsilon$ suboptimal:
  i.e., at each timestep $t$, $V^\pi_t(s) \geq V^*_t(s) - \epsilon$
  for all $s \in \mathcal{S}$,
  where $\alpha$ is $O\left( (\epsilon / |\mathcal{S}| H V^*(x_0))^2 \right)$,
  and $H$ is the horizon of the abstract MDP.
  Since the total number of transitions is bounded by $|\mathcal{S}|^2$,
  substituting for $N_{transition}$ gives the desired result.
\end{proof}

\begin{proof}[Proof of \reflem{optimality_transfer}]
  Assume that the known set is equal to the set of all abstract states at
  timestep $T$ and let $\pi$ be a policy on the abstract MDP.

  To prove the first part of \reflem{optimality_transfer},
  consider a trajectory $s_0, o_0, r_0, s_1, o_1, \cdots, o_{T - 1}, r_{T -
  1}, s_T$ rolled out by $\pi$ on the abstract MDP.
  Each abstract action $o_i$ is implemented as a subpolicy in the concrete
  MDP, so it expands to the trajectory solving the subtask of navigating from
  $s_i$ to $s_{i+1}$:
  $x_{(i, 0)}, a_{(i, 0)}, r_{(i, 0)}, x_{(i, 1)}, \cdots, o_{(i, T_i - 1)},
  r_{(i, T_i - 1)}, x_{(i + 1, 0)}$,
  where $\abstraction(x_{(i, 0)}) = s_i$,
  $\abstraction(x_{(i + 1, 0)}) = s_{i + 1}$, and
  $r_i = \sum_{j = 0}^{T_i - 1}r_{(i, j)}$,
  by definition of the abstract MDP rewards.
  Consequently, for each trajectory, $\pi$ achieves the same total reward in
  the concrete MDP as in the abstract MDP, implying
  $V^\pi_T(\abstraction(x_0)) = V^\pi(x_0)$.

  To prove the second part of \reflem{optimality_transfer},
  it suffices to show that the optimal policy on the concrete MDP achieves no
  more reward than $V^*_T(\abstraction(x_0))$, because the first part already shows that
  $V^*_T(\abstraction(x_0)) \leq V^*(x_0)$.
  Let $\pi^*$ be the optimal policy on the concrete MDP and let $\tau = x_0,
  x_1, \cdots, x_T$ be the highest-reward trajectory generated $\pi^*$
  achieving reward $R$.
  Because the known set contains all abstract states,
  in particular, it contains $\abstraction(x_T)$.
  By Assumption 1, and because $\abstraction(x_T)$ is in the known set, it is
  possible to deterministically navigate to $\abstraction(x_T)$.
  By Assumption 2, traversing to $\abstraction(x_T)$ in the abstract MDP
  achieves reward $R$.
  Hence, $V^*_T(\abstraction(x_0))$ is at least $R$, proving the desired
  result.
\end{proof}

\begin{proof}[Proof of \reflem{number_samples}]
  For the known set to cover all abstract states,
  the manager must discover all neighboring transitions for each abstract
  state and
  the worker must learn all the discovered transitions.
  The number of samples to do this is equal to the sum of:

  \begin{enumerate}
    \item
      The number of samples used by the manager to discover all transitions.
    \item
      The number of samples used by the worker to learn new transitions.
    \item
      The number of samples used by the worker to navigate to the fringes of the
      known set,
      for the worker to learn new transitions and
      for the manager to discover new transitions.
  \end{enumerate}

  \paragraph{Samples used by the manager to discover transitions.}
  At each abstract state,
  let $p$ be the probability that the manager fails to discover a particular
  abstract state on a single discovery episode.
  Let $K$ be the maximum number of outgoing transitions from an abstract state
  (maximum degree).
  Both $p$ and $K$ are polynomial in $|\Phi|$ and $|\mathcal{A}|$
  because the diameter of each abstract state is bounded by assumption.
  By setting
  the number of times the manager explores from each abstract state,
  $N_{visit} = \frac{\log{K|\Phi|} + \log{\frac{1}{\eta}}}{\log{\frac{1}{p}}}$,
  the manager finds all outgoing transitions of each abstract state
  with probability at least $1 - \eta$ by the following elementary argument.
  There are at most $K|\Phi|$ total transitions to discover,
  and the manager fails to discover each transition with probability
  $p^{N_{visit}}$.
  By the union bound,
  the probability the manager fails to discover at least one transition is
  at most $K|\Phi|p^{N_{visit}}$.
  Consequently,
  the manager explores for $O(\frac{\log{K|\Phi|} +
  \log{\frac{1}{\eta}}}{\log{\frac{1}{p}}})$
  timesteps.

  \paragraph{Samples used by the worker to learn transitions.}
  We assume that policy search with neural network function approximators can
  learn each transition at least as quickly as brute-force search over
  deterministic policies.
  By Assumption 3, the maximum number of timesteps required to traverse a
  transition $(s, s')$ is $d(s, s')$ times the diameter $H_{worker}$,
  which is at most $H = d_{max} \times H_{worker}$.
  Consequently, we bound the time required to learn each transition by
  $|\mathcal{A}|^H$, the total number of possible action trajectories for the
  worker.

  \paragraph{Samples used to navigate to the fringes of the known set.}
  The total number of samples used to navigate to the fringes of the known set
  is given by:

  \begin{equation}
    O(\sum_{s \in \Phi} N(s))
    \label{eqn:total_time}
  \end{equation}

  where $N(s)$ is the number of times the worker visits state $s$ at the
  endpoint of a transition, counted when all abstract states are in the known
  set.

  We now prove:

  \begin{equation}
    N(s) \leq N_{visit} + |E(G_s)|(|\mathcal{A}|^{H} + N_{visit}),
    \label{eqn:state_visit}
  \end{equation}

  where $E(G_s)$ is the set of transitions in the subgraph of the abstract MDP
  consisting of directed transitions from $s$.
  This holds by strong induction on $|E(G_s)|$.
  In the base case,
  when $|E(G_s)| = 0$,
  $s$ is only visited $N_{visit}$ times for the manager to explore.
  In the inductive case,
  supppose $|E(G_s)| = c$ and that the inductive hypothesis holds for all values
  less than $c$.
  $N(s)$ is at most $N_{visit} + \sum_{(s, s') \in E(G_s)}{N(s')}$.
  Since $E(G_s) = \bigcup_{(s, s') \in E(G_s)}{E(G_{s'}) \cup \{(s, s')\}}$,
  the inductive hypothesis holds for each $N(s')$.

  We bound $|E(G_s)|$ by the total number of transitions: $|\Phi|^2$.
  Substituting into \refeqn{state_visit} and \refeqn{total_time},
  yields that total time for the worker to traverse to the fringes of the known
  set is:
  $O\big(|\Phi|^3
  (|\mathcal{A}| + \frac{\log{K|\Phi|} +
  \log{\frac{1}{\eta}}}{\log{\frac{1}{p}}})
  \big)$.

  \paragraph{Total samples.}
  Adding all three terms together gives that with probability at least $1 -
  \eta$,
  the known set covers all abstract states in
  $O\big(|\Phi|^3
  (|\mathcal{A}| + \frac{\log{K|\Phi|} +
  \log{\frac{1}{\eta}}}{\log{\frac{1}{p}}}) +
  d_{max} \times H_{worker}\big)$ samples.
\end{proof}

\section{Discovering New Transitions Pseudocode}

\begin{algorithm}
  \caption{$\textsc{DiscoverTransitions}(x_0)$}\label{alg:discoverer}
  \begin{algorithmic}[1]
    \REQUIRE called at concrete state $x_0$ to discover
    transitions near $\abstraction(x_0)$
    \STATE Discovered transitions $D \gets \{\}$
    \STATE Choose $a_0 \sim \pi^\text{d}(x_0)$
    \WHILE{$n(\abstraction(x_0)) \leq N_{visit}$}
      \FOR{$t = 1$ to $T_\text{d}$}
        \STATE Observe $x_t, r_t$
        \STATE Add transition $(\abstraction(x_{t - 1}), \abstraction(x_t))$
        to $D$
\STATE Choose $a_t \sim \pi^\text{d}(x_{1:t}, a_{1:t-1})$
      \ENDFOR
      \STATE Continue exploring: reset $x_0 \gets x_{T_\text{d}}$ and choose $a_0 \sim \pi^\text{d}(x_0)$
    \ENDWHILE
    \STATE Return $D$
  \end{algorithmic}
\end{algorithm}
 
\end{document}